# LLM-based Text Simplification and its Effect on User Comprehension and Cognitive Load

**Authors**
Theo Guidroz*, Diego Ardila*, Jimmy Li*, Adam Mansour, Paul Jhun, Nina Gonzalez, Xiang Ji, Mike Sanchez, Sujay Kakarmath, Mathias MJ Bellaiche, Miguel Ángel Garrido, Faruk Ahmed, Divyansh Choudhary, Jay Hartford, Chenwei Xu, Henry Javier Serrano Echeverria, Yifan Wang, Jeff Shaffer, Eric (Yifan) Cao, Yossi Matias, Avinatan Hassidim, Dale R Webster, Yun Liu, Sho Fujiwara, Peggy Bui, Quang Duong

*Equal contributions

**Affiliation**
Google

**Correspondence**: liuyun@google.com

**Abstract**
Information on the web often surpasses users' reading level, with knowledge resources like scientific publications, Wikipedia, and news sites regularly exceeding the reading level of the general population. To help address this, we used a self-refinement approach to develop a large language model (LLM) capability for minimally lossy text simplification. To validate our approach, we conducted a randomized comprehension study involving 4563 participants and 31 texts spanning 6 broad subject areas: PubMed (biomedical scientific articles), biology, law, finance, literature/philosophy, and aerospace/computer science. Participants were randomized to viewing original or simplified texts in a particular subject area, and subsequently answered multiple-choice questions (MCQs) that tested their comprehension of the text. The participants were also asked to provide qualitative feedback about the text's difficulty. Our results indicate that participants who read the simplified text answered more MCQs correctly than their counterparts who read the original text (3.9% absolute increase, $p<0.05$). This gain was most striking with PubMed (14.6%), while more moderate gains were observed for finance (5.5%), aerospace/computer science (3.8%) domains, and legal (3.5%). Notably, the results were robust to whether participants could refer back to the text (original or simplified) while answering MCQs. The absolute accuracy decreased by up to ~9% for both original and simplified setups where participants could not refer back to the text, but the ~4% overall improvement in MCQ accuracy persisted. Finally, participants' self-reported perceived ease based on a simplified NASA Task Load Index was greater for those who read the simplified text (absolute change on a 5-point scale 0.33, $p<0.05$). This randomized study, involving an order of magnitude more participants than prior works, demonstrates the potential of LLMs to make complex information easier to understand. Our work aims to enable a broader audience to better learn and make use of expert knowledge available on the web, improving the accessibility of information.

# Introduction

In the digital age, expert information has become widely accessible, but much of it remains too difficult for the general populace to understand. Scientific publications, encyclopedic resources like Wikipedia, and articles from news outlets are often written at a level that exceeds the recommended reading level for broad audiences.[1,2] In the United States, for instance, over half of adults have a reading level below that of the 6th grade[3], while much of the information available on complex topics such as health surpasses this threshold[4]. This disconnect poses a critical challenge to effective information dissemination, potentially leading to misinterpretations, reduced engagement, and ultimately, a diminished capacity for informed decision-making. The implications are particularly pronounced in domains such as healthcare, law, and finance, where misinterpretations can have substantial real-world repercussions.

Recognizing the importance of making information not just accessible but also comprehensible, various initiatives have been undertaken to mitigate this challenge. Governmental bodies, for example, have implemented regulations mandating that public-facing documents adhere to specific readability standards,[5] and have provided guidance on how to produce readable content.[2] However, these efforts are often hindered by the inherent limitations of traditional linguistic metrics used to assess reading level. Established measures like the Flesch-Kincaid Grade Level and the SMOG index, while widely utilized, are known to exhibit inconsistencies and fail to capture the nuance of what makes text difficult to comprehend.[6,7] These metrics typically rely on surface-level features such as sentence length and word or syllable frequency, neglecting deeper semantic and contextual factors that significantly influence reader understanding. Moreover, they may not adequately account for the domain-specific jargon and technical terminology that characterize expert discourse. Consequently, a more sophisticated and adaptable approach is needed to bridge the gap between complex information and broad audiences.

In this study, we explore the potential of using large language models (LLMs) – specifically Gemini[8] – to achieve minimally-lossy text simplification. Our methodology involves developing automated evaluation techniques for (1) measuring how readable the rewritten text is compared to the original, and (2) measuring how faithfully the rewritten text captures the same informational content as the original. Using automated evaluation techniques for these aspects, we further leverage Gemini to analyze performance and iteratively refine prompts and few-shot examples. To validate our approach, we conducted a randomized controlled study. Participants were assigned to different study arms, where they read original or simplified texts across a range of domains, including PubMed, biology, law, finance, literature/philosophy, and aerospace/computer science. Subsequently, they answered multiple-choice questions (MCQs) designed to assess their comprehension of the presented information. Participants then provided subjective evaluations of their confidence and their cognitive workload using a simplified NASA Task Load Index. Our findings demonstrate that participants who reviewed simplified texts achieved significantly higher accuracy on the MCQs, were more confident in their answers, and reported a substantial improvement in reported ease of the task. These results underscore the efficacy of LLM-driven text simplification as a means of enhancing information accessibility without compromising the integrity of expert knowledge.

## Related work

In earlier literature, *text simplification* was proposed to reduce the complexity of text, primarily through syntactic simplification (simplification of sentence structure) and/or *lexical simplification* (replacement of complex words with simpler variants without syntactic edits).[9–11] Text simplification has since been recognized to have broad potential applications, including aiding individuals with language disabilities, language learners, non-native speakers, users of small-screen devices. Methodologically, recent work often involves training sequence-to-sequence models (eg, BART) on parallel corpora, such as medical abstracts paired with their Plain Language Summaries (PLS), incorporating penalties for complex terminology or using reinforcement learning with rewards tailored for relevance and simplicity.[12,13] A key challenge, especially in technical domains, is providing necessary background context; works such as Retrieval-Augmented Lay Language (RALL) generation address this by integrating external knowledge (e.g., definitions from UMLS or Wikipedia) during simplification.[14] Approaches like "Paper Plain" highlight a user-centric approach, developing interfaces that go beyond simple text transformation to include features like question-based navigation, section summaries ("gists"), and term definitions, while acknowledging the critical need to manage potential factual inaccuracies ("hallucinations") in generated content.[15]

To quantify simplification, metrics borrowed from machine translation (eg, BLEU, ROUGE[16,17]) and specific simplification metrics (eg, SARI, FKBLEU, LENS[18,19]) have been proposed to assess semantic preservation and simplicity, though they often require gold-standard reference texts and show limited correlation with human judgments on aspects beyond grammar and adequacy.[19] Consequently, research has increasingly focused on alternative evaluation methods, such as training language models (like BERT/SciBERT) to discriminate between complex and simple styles,[20] developing metrics focused on factuality to detect meaning changes (insertions, deletions, substitutions)[21] based on tools such as entailment checkers,[22] and relying on comprehensive human evaluation assessing dimensions like fluency, coherence, simplicity, understandability, and factual correctness.[12,21] Alternatively, works such as Martínez et al.[23] and Carrer et al.[24] also use professional easy-reading rewriters to qualitatively judge simplified text based on easy-reading criteria.

While works have explored various technical approaches to simplification, including both automated and manual evaluation of the simplified content, rigorous studies (particularly randomized studies) evaluating the effects of text simplification on reader comprehension are less common. Agrawal et al. explored text simplification with 112 participants, with each participant seeing 6 text excerpts and answering 3 MCQs;[25] Kreijkes et al. explored LLM based note-taking by 405 students and found that, qualitatively, students appreciated the LLM for reducing complexity[26]; Crossley et al. explored simplified texts with 48 participants [27]; Säuberli et al., Carrer et al. conducted text comprehensibility studies for German text for a target group of 18 people with intellectual disabilities and a control group of 18[24,28]; Rello et al. performed a similar study with 23 dyslexic individuals and 23 controls for Spanish text.[29]

## Methods

We developed an automated iterative prompt refinement system for minimally lossy text simplification using several Gemini LLMs. Our system has several components: (1) the text simplification model, (2) automated evaluation (autoeval) models, (3) a ranking module, and (4) a prompt refinement model (Figure 1A).

This system was developed using a text dataset (n=123) with a mean length of 61.7 words (SD: 31.3). The text simplification model is based on Gemini 1.5 Flash with a custom prompt and few-shot examples drawn from the dataset, both of which were iteratively improved.

## Autoeval system

The 2 autoeval models targeted readability and fidelity. The readability model was developed because of well-appreciated limitations of Flesch-Kincaid and other linguistic-based readability scoring systems,[6,7] and comparisons showing limited correlation between a delta in the Flesch-Kincaid score and human readability scoring of ease of comprehension. The readability model is a Gemini Ultra model with a custom prompt, and was developed to take a single text as input, and output an integer in [1,10]. The argmax (ie, integer) was taken to be the eval rating, and the log-likelihood taken to be the confidence level. Guided by human readability evaluations on 129 original/simplified text pairs, the prompt was iteratively improved via a feedback loop that improved the prompt using previously misclassified examples.

The fidelity autoeval uses Gemini 1.5 Pro with a custom step-by-step reasoning prompt to understand completeness and entailment. Completeness measures whether a rewritten text contains all information in the original and entailment measures whether a rewritten text logically follows from the original. The prompt contains instructions to compare the original and rewritten texts, by first breaking up the original text into atomic claims and then mapping each of the claims to excerpts in the rewritten text, and listing any claims in the rewritten text that do not map to any claim from the original text. Then, the claims are analyzed for 8 errors in 3 categories (below), with weights (in square braces below) for error scoring:

1. Information loss (entirety of the claim missing [2], specificity lost [1], nuance/connotation lost [2])
2. Information gain (unfactual [4], off topic [1])
3. Distortion (loss in factuality [4], significant loss in fidelity [3], minor loss in fidelity [1])

Categories 1 and 3 were considered completeness errors, while categories 2 and 3 were considered entailment errors. The final overall error score was a weighted sum based on each error's weight, and divided by 10 to bring it into a similar numeric range as the readability model's output. These weights were determined empirically based on subjective judgement on the relative severity of each type of error.

## Prompt refinement system

To refine the prompt, a Gemini 1.5 Pro model was leveraged to iteratively refine the text simplification model, starting from a 'seed' prompt. In each iteration, the current prompt is used to simplify text in the dataset, and the autoeval models (described above) are run across the dataset. The prompt's score was defined as the averaged readability score minus the averaged error scores, and the best prompts were tracked based on this score. A Gemini model was instructed to improve the prompt in each iteration and the final prompt was selected after manually terminating the iteration loop (Results).

## Randomized study

To study the effect of using a minimally lossy text simplification model, we ran a randomized survey based study in Qualtrics (Provo, UT), where consented participants read text with and without simplification, and answered several questions.

### Study dataset

We curated 31 text excerpts spanning 6 broad subject areas (Table 1): PubMed (biomedical scientific articles), biology, law, finance, literature/philosophy, and aerospace/computer science. The texts were extracted from various publicly accessible sources including PubMed Central, Project Gutenberg, WikiNews, and more (Table 1). Sources were selected to encompass a broad range of topics that we expected people to be browsing on the Web. We also prioritized content that we expected to be challenging for people but for which some people may be trying to understand (eg, because they know someone who just got diagnosed with a condition, are signing a lease, curious about the latest famous scientific prize awardee in the news, etc). Text excerpts were manually extracted and categorized into the aforementioned 6 broad topic areas. For each (original) text excerpt, 2 multiple choice questions (MCQs) were written, with at least 2 researchers reviewing each question and answer before finalization. In a small portion of texts, Gemini was used for inspiration on the question and options; all questions were edited and inspected manually prior to finalization. This process was completed while blinded to the text simplification model outputs for this dataset.

### Study design

We employed a randomized complete block design (RCBD) using five experimental arms and one control arm (Figure 1), structured to assess the impact of simplification:

- "Open-book" setup, where participants could see the text(s) relevant to that study arm, at the same time as the questions
    - Arm 1 / Control (original text, open-book)
    - Arm 2 (simplified text, open-book)
    - Arm 3 (both original and simplified texts, open-book)
- "Closed-book" setup, where participants were shown the text(s) relevant to that study arm, but could not refer back to the text(s) when viewing or answering the questions
    - Arm 4 / Control (original text, closed-book)
    - Arm 5 (simplified text, closed-book)
    - Arm 6 (both original and simplified texts, closed-book)

Participants in the closed-book condition received instructions emphasizing that they could not return to the text once they proceeded to the questions.

### Participant recruitment and sample size

We recruited a total of 4,563 consented participants across the 6 topic areas (Table 2). For each topic area, participants were randomly assigned to each of the six experimental arms. Due to the manual work required to select text and generate MCQ questions, we optimized for the number of total responses by increasing the number of participants instead of the number of texts or questions.

We targeted enrollment based on demographic variables to approximate the US Census age distribution after adjusting for removal of under-18s (see below): 16% 18-24 years, 17% 25-34 years, 16% 35-44 years, 16% 45-54, 35% 55+ years old. We also targeted an equal ratio of man:woman.

### Participant flow

Participants self-reported demographic details including whether they were native English speakers, their confidence in understanding complex written English materials, their age, gender, education level, and domain/background. For all study arms, participants meeting any of the following were screened out: (1) not a native speaker and does not use English as a primary language, (2) below 18 or the age of majority in their state, (3) did not consent. For each topic area, participants were also screened out if they self-reported educational background in the topic area of that study arm.

All participants then proceeded to the texts in that topic area (Table 1) and associated questions. Each text was followed by two 5-option multiple-choice questions (MCQs) designed to assess comprehension of the text. Each MCQ was followed by a confidence scale. Each text section (the text, 2 MCQs, and 2 confidence questions) was followed by a simplified NASA Task Load Index[30] (cognitive load) question. Both the confidence and cognitive load questions were bipolar on a 5-point scale, and (mapped to integers in [-2, 2] for analysis). For confidence the question was “How confident are you in your response to the previous question?” with answer options: Very confident, Somewhat confident, Neither confident nor unconfident, Somewhat unconfident, and Very unconfident; for cognitive load the question was “How easy or difficult did you find it to answer the previous 2 questions, based on your understanding of the text you read?”, with answer options: Very easy, Somewhat easy, Neither easy nor difficult, Somewhat difficult, and Very difficult.

Finally, at the end of the survey, participants answered a simple attention check question; responses from participants who provided nonsensical answers to this question were discarded.

### Pilot studies

The study described above reflects the study executed and run after a series of 3 conceptually similar but smaller and simpler pilot studies, from which qualitative and quantitative results guided model development and final study design. We summarize the main qualitative insights here. First, the use of long texts may change the participants’ task from one of comprehension to information retrieval (this pilot was done only in the “open-book” condition); therefore we reduced the length of the text to enable a more controlled test. Second, free text response answers as a measure of comprehension were not a reliable measure of comprehension in this study population, possibly due to survey platform users being incentivized to answer questions quickly. For example, we observed that many free-text answers were too brief to be useful for measuring participant comprehension. Third, prior expertise in a given area may confound results by encouraging participants to rely on their own knowledge to answer questions.

### Statistical analysis

We performed statistical analyses using Python, leveraging the NumPy, pandas, statsmodels, SciPy libraries. For each participant, we calculated the proportion of correct MCQ answers. Individual-level accuracies were calculated from the per-question data. We then aggregated these individual metrics to the group level, based on the experimental conditions (open-book vs. closed-book and original vs. simplified vs. original plus simplified).

We used linear regression to analyze the effect of simplification on accuracy of MCQ responses. For each experimental condition, we calculated the difference in mean accuracy between the simplified text group and the original text group. This analysis is statistically equivalent to a two-sample t-test.

We used the matplotlib and seaborn libraries to visualize the data. Bar plots were used to display mean accuracy and Likert values across experimental conditions and question themes, with error bars representing confidence intervals. Scatter plots were used to compare original vs. simplified text accuracy at the question level.

# Results

Figure 1 summarizes our automated approach to a minimally lossy text simplification model (Methods), involving Gemini-based models for both autoeval and iterative refinement of the text simplification model. The iterative process was terminated after 824 iterations, when the improvements plateaued. The final prompt was manually selected from the best performing prompts, and 4 examples of simplifying short inputs were added.

We then tested the text simplification model in a randomized study with 4,563 consented participants and texts spanning 6 topic areas (Methods, Tables 1 and 2). Altogether, the participants provided a total of 49,582 MCQ answers.

Overall, we found a 3.9% improvement ($p<0.05$, 95% CI = (1.6, 6.3), from 44.3% to 48.2%) in MCQ accuracy when participants saw the simplified text, compared to the original text (Figure 3, left). This improvement varied across topic areas, with the largest improvement for PubMed (14.6% with 95% CI = (8.7, 20.6)), and more modest improvements for other areas such as finance, aerospace/computer science, and legal. These results were robust to a setting where participants could not see the text (original or simplified) when answering questions (Figure 3, right).

The average improvements for participants' self-reported confidence per MCQ question (converted to a [-2, 2] scale, see Methods) was 0.24 ($p<0.05$, 95% CI = (0.21, 0.27)). Similarly, the confidence improvements varied by topic area, though unlike that for MCQ accuracy, for confidence we observed significant improvements for 4 or 5 topic areas (PubMed, Literature/Philosophy, Finance, Biology, and Law), with the final topic (Aerospace/Computer Science) trending positive. These results were similarly robust to the setting where participants could not see the text while answering questions, though the confidence improvements were generally more modest. In terms of participants' self-reported ease of reading the text and answering the 2 MCQs (on a per text basis), we observed a significant improvement of 0.33 ($p<0.05$, 95% CI = (0.28, 0.38)), with the same 4 of 5 topics (as the confidence results) being significant (Figure 5).

When plotting the simplified vs. original accuracy on a per question basis (Figure 6 A and B), we observe that there is a trend towards greater accuracy improvements for MCQs where the original accuracy was lower (eg, below 0.4). Because many PubMed MCQs' original accuracies were low, this trend towards greater assistance for harder questions was also more pronounced for the PubMed topic area (Figure 6 C and D).

The top 3 MCQs with the greatest change in accuracy with the simplified text, were all from the PubMed topic area, with absolute improvements of 38%, 28%, and 28%, respectively (Table 3). Supplementary Data lists all texts, rewrites, and MCQs.

## Discussion

In this study, we developed and evaluated a Gemini-based system for minimally lossy text simplification, designed to enhance comprehension of complex or technical information available on the Web. Our automated approach to prompt iteration and autoeval showcases the use of Gemini at all stages of the development process including: automated evaluation with complex reasoning chains, analysis of successful and unsuccessful rewrites, and prompt improvement including few-shot example selection. Our approach should be generalizable to other LLM applications, and rendered more efficient with time as the capabilities of base models improve further (improving each step of our pipeline).

Our randomized controlled comprehension study, involving almost 5,000 participants and 50,000 MCQ answers across 6 topic areas, is also notable for its size, roughly 1 order of magnitude larger in size of participant pool than prior work. We observed a statistically significant 4% absolute improvement in MCQ accuracy when participants reviewed simplified texts. Notably, the most substantial gains were observed in PubMed (15%), with more moderate but significant improvements in legal, finance, and technical (aerospace / computer science) domains. The observed benefits were robust to a setting where participants could not access the source text when answering questions, highlighting its ability to help both comprehension and short term retention. The strong improvements in self-reported confidence and perceived ease of this task reflect another strength of our approach to text simplification: making it easier for people to understand complex information.

The results above highlight a curious but perhaps not unexpected juxtaposition: while MCQ accuracy gains were modest for many topic areas, the confidence and task ease saw significant improvements. First, this highlights that confidence (improvement) is not a good proxy for comprehension and emphasizes the need to measure both comprehension and confidence. Second, regardless of the accuracy trends, participants found the task (reading the technical material and answering questions) to be easier with simplification, suggesting the usefulness of text simplification while browsing the web.

The benefits of text simplification could be substantial by bridging the gap between complex, expert-written information and the relatively more limited domain-specific comprehension abilities of a non-expert audience. As such, this approach can facilitate better access to the world's information. Our specific results suggest that comprehension could improve for complex medical information in PubMed, a public resource for researchers and patients alike. More modest benefits in areas like finance could help improve financial literacy for many.

Our mechanism of text simplification is likely most useful to users who run into challenging text when viewing content on the web. They could then manually highlight a section of text and retrieve the simplified version. This ensures that the simplified text can be read within the context of the original source. Our model leverages an extremely fast model, Gemini 1.5 Flash, to ensure fast simplification. Alternatively, an automated model (perhaps personalized to each individual) could look for and highlight challenging text sections, enabling users to trigger simplification on-demand. This latter automated approach trades off increased computation (to detect and pre-generate simplified text) with the potential to eliminate perceived latency, by running the simplification model prior to the user reading that section. In either case, our simplification approach could seamlessly help users understand complex material that they encounter on the web.

Despite the positive and statistically significant improvements seen above, our study has limitations. First, we leveraged a survey platform where participants reflect a broad, non-expert population who are not specifically looking to understand the material presented. Moreover, the monetary aspects of such a platform could have incentivized participants to answer questions quickly instead of spending significant time to optimize for accuracy. Though the randomization ensures that the measured effects controls for these factors and preserves the integrity of the study, these factors likely depressed measured accuracies in all study arms. The specific comprehension impact on motivated users who want to understand the material, will also need to be studied in future work. Secondly, while our LLM aimed for lossless simplification, errors in the simplification process could have introduced inaccuracies or ambiguities. Whether the poorer MCQ accuracies post-simplification (dots below the diagonal in Figure 6) reflect random variation or information distortion will need to be studied with a larger sample size. Thirdly, the MCQs, while designed to assess comprehension in an easy-to-score way, may not have been sufficiently challenging to capture the full depth of understanding. Future research could further explore alternative testing methodologies, such as open-ended questions (“explain what you learned”) to provide a more nuanced assessment of comprehension. Per our pilot studies however, such an effort will need to engage a motivated audience who are incentivized to learn the material well and provide detailed responses. Finally, our approach as it is implemented currently relies on the knowledge encoded within the LLM, and consequently, may struggle with novel terms or recent slang. Future approaches leveraging a live web search to inform the simplification model, may help address any recency information gaps.

In conclusion, our study provides compelling evidence that LLM-based minimally lossy text simplification can significantly enhance comprehension of expert information. However, it is important to recognize that this approach is not a panacea for understanding. While it can effectively bridge the gap between complex language and comprehension, it is essential to consider the limitations of the technology and to explore complementary strategies for improving information accessibility. Future research should focus on further refining the simplification process, developing more robust evaluation methodologies, and addressing the limitations highlighted in this study.

## Acknowledgements

Our appreciation goes to Mike Schaekermann and Michael Howell for helpful feedback on the manuscript. This paper was drafted manually and some subsections were refined using Gemini 2.0 Flash before undergoing further manual iterations. All numerical and statistical analyses were done manually. The authors take all responsibility for the accuracy and contents of the paper.

## Figures & Tables

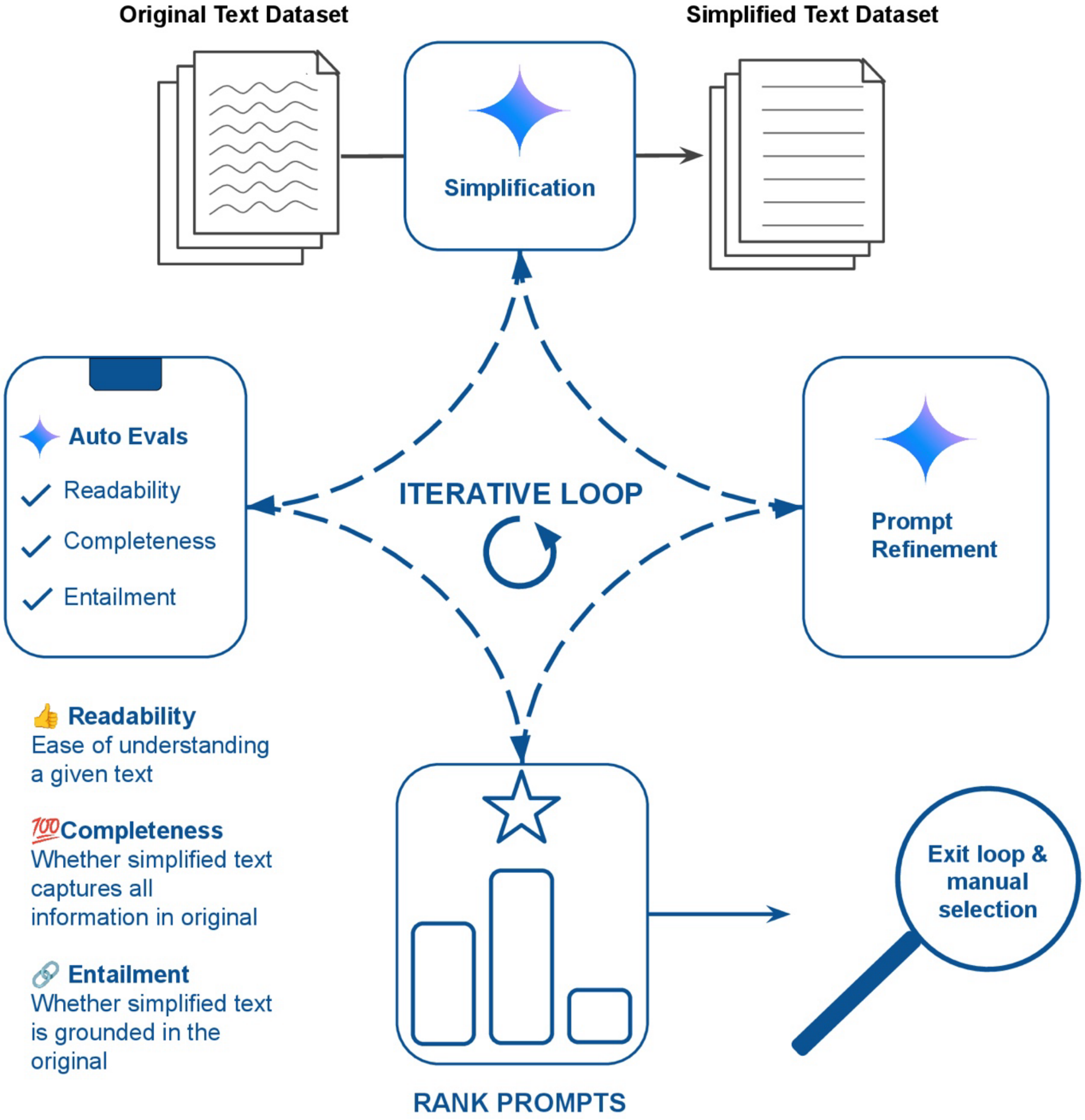

**Figure 1: Summary of Gemini-based approach for minimally lossy text simplification.** Our goal is to develop a model that simplifies text while avoiding either adding or losing information (top, simplification model). This involved creating automated evaluations ("autoevals") for readability and information fidelity (completeness and entailment), and iteratively using the autoevals to rate the candidate simplification model, and improve the prompt. All models used (indicated by the star) are Gemini based.

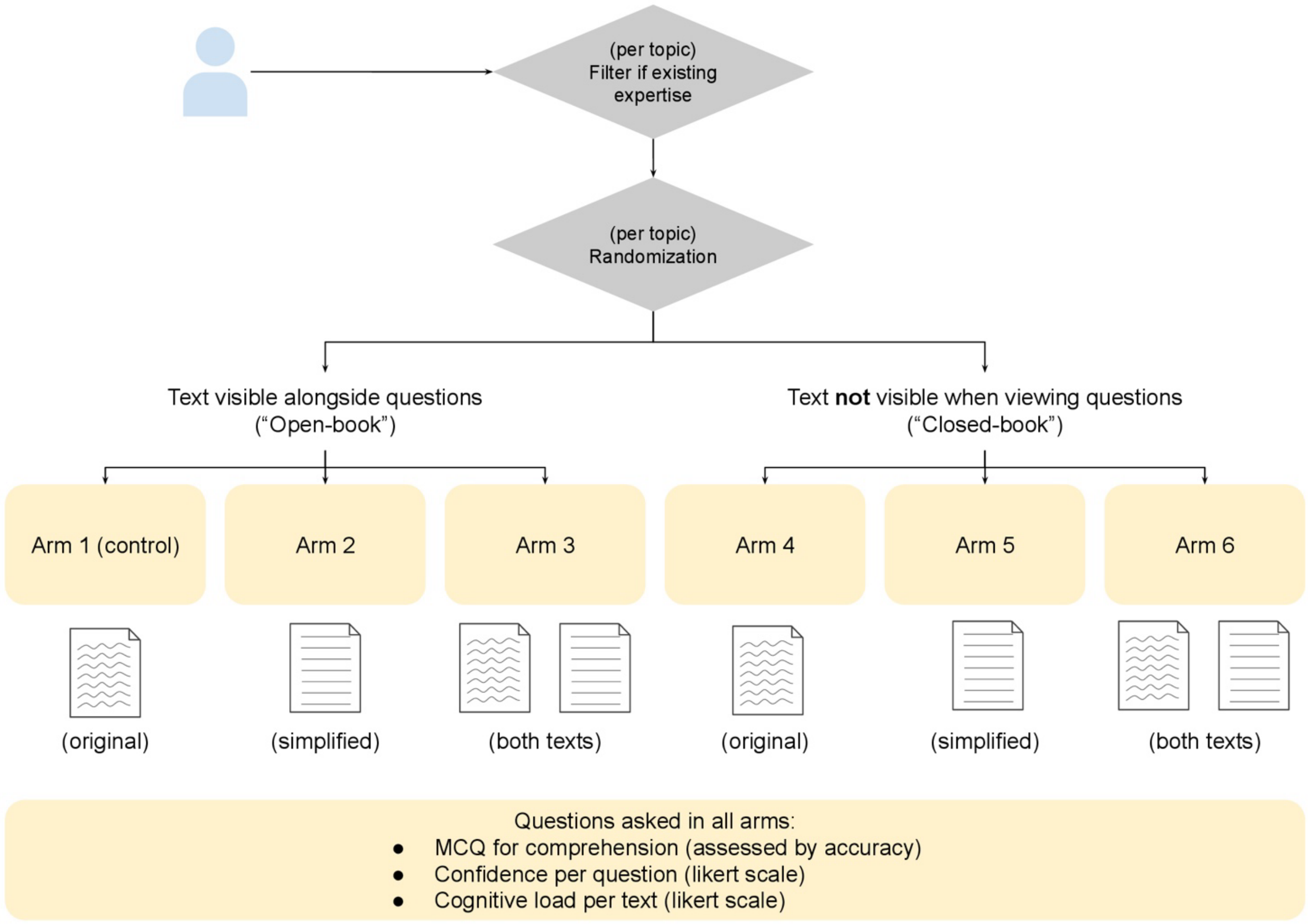

**Figure 2: Study design to evaluate the simplification model with real texts.** For each of the 6 topic areas (Table 1), participants are filtered for existing expertise in that topic area, and consented (Methods). Participants were then randomized to one of 6 study arms, where they view the original and/or simplified texts, and answer multiple choice questions (MCQs) and rate their confidence level per MCQ as well as a simplified NASA Task Load Index ("cognitive load" or difficulty/ease) of that question.

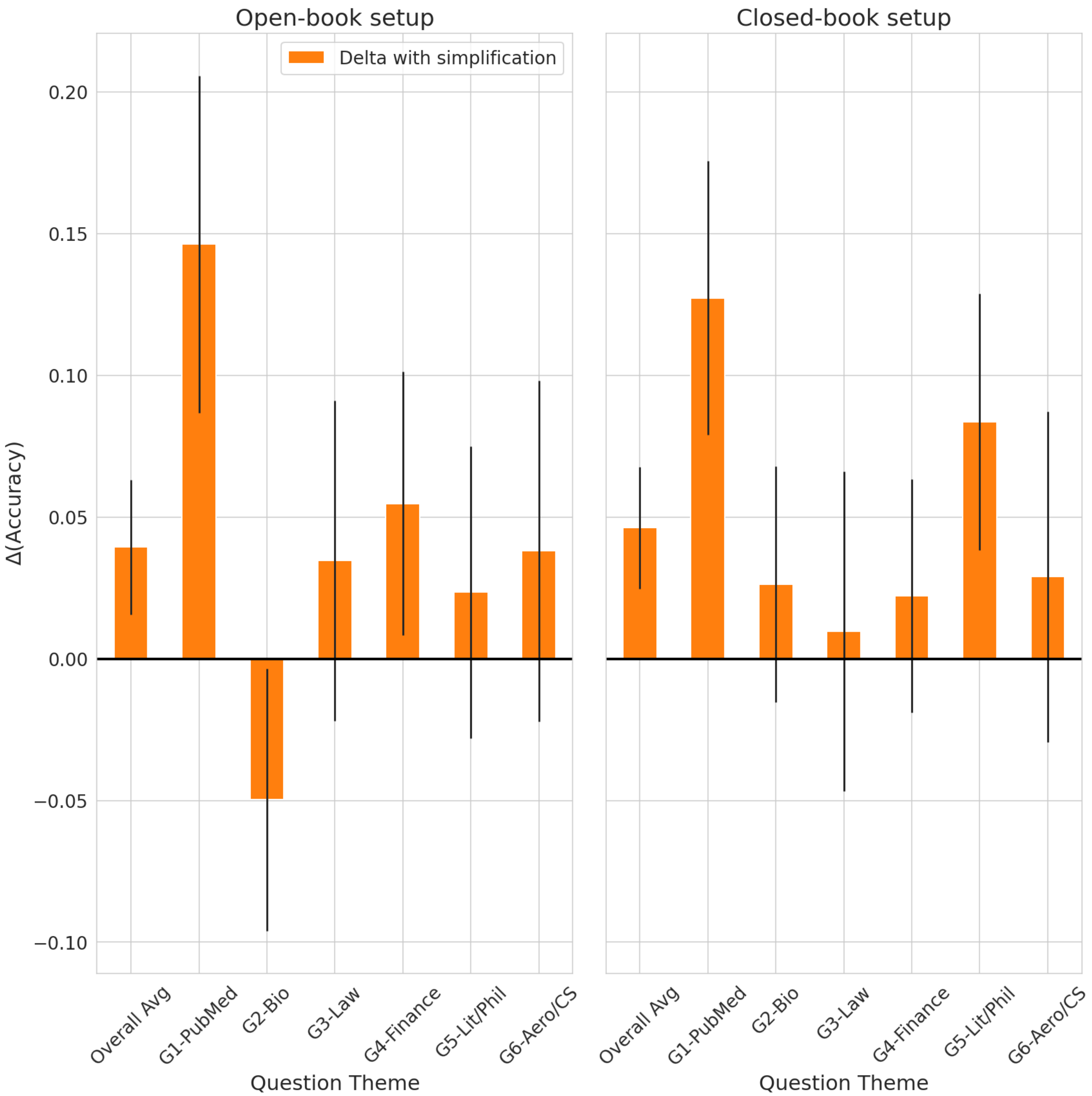

**Figure 3: Change in multiple-choice question (MCQ) accuracy when participants viewed the original vs simplified texts.** The left bars indicate the "open-book" setup where participants could see the associated text while answering the questions (arm 2 minus arm 1, see Figure 1), while the right bars indicate the "closed-book" setup where participants could not (arm 5 minus arm 4, see Figure 2). The leftmost bar in each setup indicates the overall change across all questions; the other 6 bars indicate the per-topic changes. Positive values indicate improvements in MCQ accuracy with simplified texts.

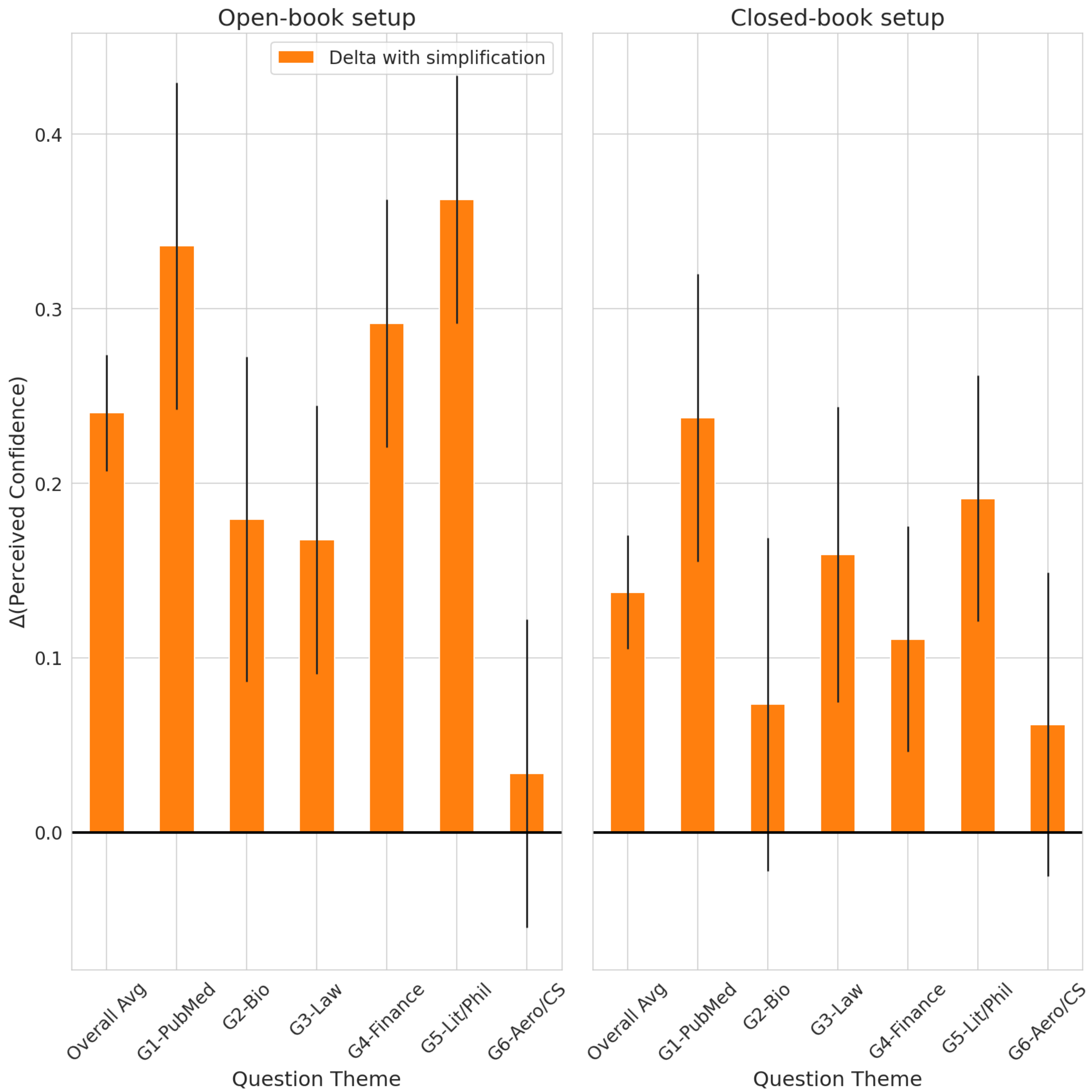

**Figure 4: Change in perceived confidence per multiple-choice question (MCQ) when participants viewed the original vs simplified texts.** Plot setup mirrors that of Figure 3; positive values indicate improvements in survey participant MCQ answer confidence with simplified texts.

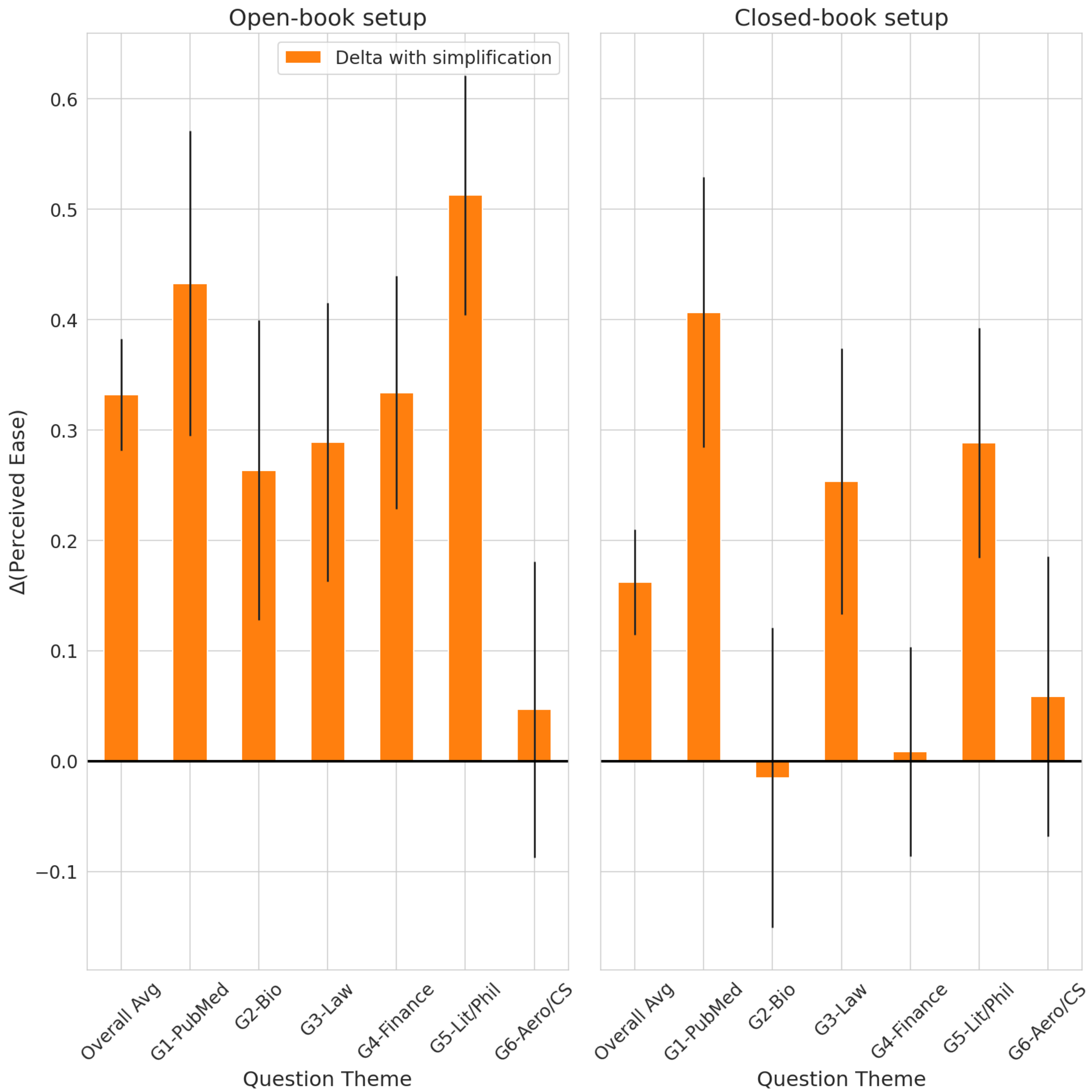

**Figure 5: Change in perceived ease per text when participants viewed the original vs simplified texts.** Plot setup mirrors that of Figure 3; positive values indicate improvements in survey participant self-reported ease of the task, with simplified texts.

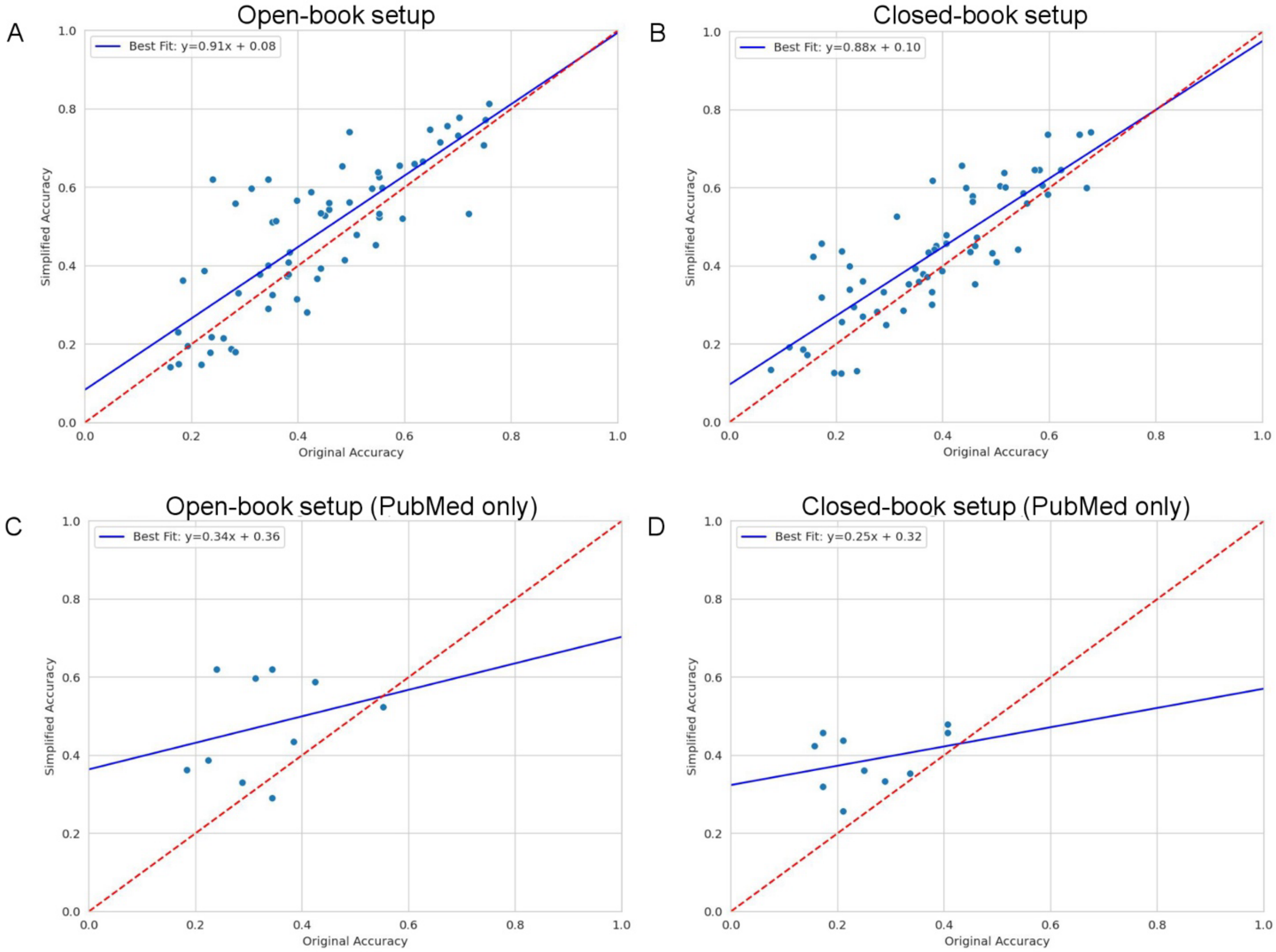

**Figure 6: Comparison of per-question accuracy when participants viewed the original vs. simplified texts, across all questions (A,B) and specifically for the PubMed category where the accuracy gains were most obvious (C,D).** (A,C) represent the “open-book” condition where participants could view the associated text while answering the question, while (B,D) represent the “closed-book” condition where participants could not view the associated text while answering the question. Red dotted lines indicate the diagonal (dots above have higher accuracies with simplified texts). The blue line indicates the linear line of best fit, with a clockwise rotation from the diagonal indicating that questions for which the original text’s accuracy was lower yielded a higher absolute improvement when the text was simplified.

**Table 1: Categories of texts used in study.**
All texts, MCQs, and answers are listed in Supplementary Data.

| Category | Description | Exclusion of participations with self-reported expertise in: | No. texts (No. questions) | No. participants (No. responses) |
|---|---|---|---|---|
| PubMed | PubMed scientific articles | Biology, life sciences, healthcare/medicine | 5 (10) | 783 (7,830) |
| Bio | Biological articles and news | Biology, life sciences, healthcare/medicine | 4 (8) | 818 (6,544) |
| Law | Legal documents (eg, housing, contracts) | Law | 5 (10) | 806 (8,060) |
| Finance | Personal finance (eg, tax) and economics (eg, employment) | Finance/economics/accounting | 7 (14) | 772 (10,808) |
| Lit/Phil | English literature | Literature / philosophy | 6 (12) | 817 (9,804) |
| Aero/CS | News related to aerospace, computer science, robotics | Aerospace, computer science, robotics | 4 (8) | 817 (6,536) |

**Table 2: Self-reported participant characteristics.**

| Characteristic | | Count (%) |
|---|---|---|
| Gender | Man | 2,313 (48%) |
| | Woman | 2,462 (51%) |
| | Nonbinary / prefer not to answer | 38 (1%) |
| Age | 18 to 24 | 645 (13%) |
| | 25 to 34 | 817 (17%) |
| | 35 to 44 | 768 (16%) |
| | 45 to 54 | 784 (16%) |
| | 55 to 64 | 750 (16%) |
| | 65 to 74 | 762 (16%) |
| | 75 or over | 281 (6%) |
| | Prefer not to answer | 6 (0%) |
| English proficiency | Native speaker | 4,493 (93%) |
| Confidence in english | Extremely confident | 3,181 (66%) |
| | Very confident | 965 (20%) |
| | Moderately / Slightly / Not at all confident | 667 (14%) |
| Education | High school diploma or equivalent | 1,253 (26%) |
| | College credit without degree | 1,190 (25%) |
| | College | 1,705 (35%) |
| | Graduate or professional | 536 (11%) |
| | Less than high school / Prefer not to answer | 129 (3%) |

**Table 3: Qualitative examples of the 3 multiple choice questions (MCQs) and associated original and simplified texts with the greatest improvement in MCQ accuracy with text simplification.** Bolding indicates the section of the original/simplified text that the question relates to. Each of the excerpts are from articles available from PubMed Central (PMCIDs PMC10177208, PMC11475612, PMC10363436); the articles are each licensed by their respective authors under the terms and conditions of the Creative Commons Attribution 4.0 International Public License.

| Improvement rank (% absolute change) | Original Text | Simplified Text | Question |
|---|---|---|---|
| 1 (38%) | Combined pulmonary fibrosis and emphysema (CPFE) represents an increasingly recognized, progressive lung pathophysiology. The disease is associated with strong male predominance, a lack of specific treatment options and poor prognosis. The complex pathology of this condition involves emphysematous destruction of lung parenchyma, diffuse interstitial fibrosis, changes in the composition of lung immune cells, increased production of immunomodulatory factors and the prominent remodeling of pulmonary vasculature. Such an existence of obstructive and restrictive changes in the same lungs results in unexpectedly preserved lung volumes, while in contrast gas exchange is impaired. | Combined pulmonary fibrosis and emphysema (CPFE), a lung condition that's becoming more common, gets worse over time. It mostly affects men, has limited treatment options, and a poor outlook. This complex condition involves damage to the lung tissue from emphysema, a disease that damages the air sacs in the lungs, and widespread scarring of the lung tissue, called fibrosis. The immune cells in the lungs change, and the body makes more immunomodulatory factors, substances that control the immune system. The blood vessels in the lungs also change a lot. Since the lungs have both obstructive and restrictive problems, the lung volume stays about the same, which is unusual. However, the lungs cannot exchange gases properly. | According to the passage, what happens to the lungs in people with CPFE? |
| 2 (28%) | The liver presents a well-organized cellular architecture, which mirrors its complex and wide variety of functions. The liver is typified by the parenchymal cells (hepatocytes) and non-parenchymal cells. The latter include resident macrophages (Kupffer cells, KCs), hepatic stellate cells (HSC), lipocytes cells, and the sinusoidal intrahepatic lymphocytes (IHL). A signaling network connects parenchymal and non-parenchymal cells, maintaining hepatic functions under homeostatic conditions. Alterations in hepatic and organ-related metabolic processes and pathways lead to changes in the hepatic histological spectrum, progressing toward chronic disease, which is accompanied by metabolically altered hepatocytes, inflammation, and fibrosis. | The liver has a very organized structure, which reflects its many complex jobs. The liver is made up of two main types of cells: parenchymal cells (hepatocytes, which are liver cells) and non-parenchymal cells. Non-parenchymal cells include Kupffer cells (KCs), which are a type of macrophage (a cell that eats up harmful things), hepatic stellate cells (HSC), lipocytes (fat-storing cells), and sinusoidal intrahepatic lymphocytes (IHL, a type of white blood cell in the liver). A network of signals connects the parenchymal and non-parenchymal cells, keeping the liver working properly under normal conditions. When the liver's and body's metabolic processes, or the ways the body uses energy, change, it can lead to changes in the liver's structure, which can lead to chronic liver disease. Chronic liver disease is accompanied by changes in liver cells, inflammation, and fibrosis (scarring). | According to the passage, what is one of the main causes of chronic liver disease? |
| 3 (28%) | In the intermittent access model, we examined the ability of semaglutide to decrease alcohol intake and block relapse-like drinking, as well as imaging the binding of fluorescently marked semaglutide to nucleus accumbens (NAc) in both male and female rats. The suppressive effect of semaglutide on alcohol-induced locomotor stimulation and in vivo dopamine release in NAc was tested in male mice. We evaluated effect of semaglutide on the in vivo release of dopamine metabolites (DOPAC and HVA) and gene expression of enzymes metabolising dopamine (MAOA and COMT) in male mice. | In the intermittent access model, we studied how semaglutide, a medicine, could reduce alcohol use and prevent a return to drinking in rats, both male and female. We also looked at how fluorescently marked semaglutide attached to the nucleus accumbens (NAc), a part of the brain involved in reward and motivation. In male mice, we tested how semaglutide affected movement caused by alcohol and dopamine release in the NAc. We also checked how semaglutide influenced the release of dopamine metabolites (DOPAC and HVA), chemicals that are created when dopamine is broken down, and the activity of genes that control enzymes (MAOA and COMT) involved in breaking down dopamine in male mice. | According to the passage, what is one way the researchers measured the effects of semaglutide on the male mice? |

# Supplementary Information

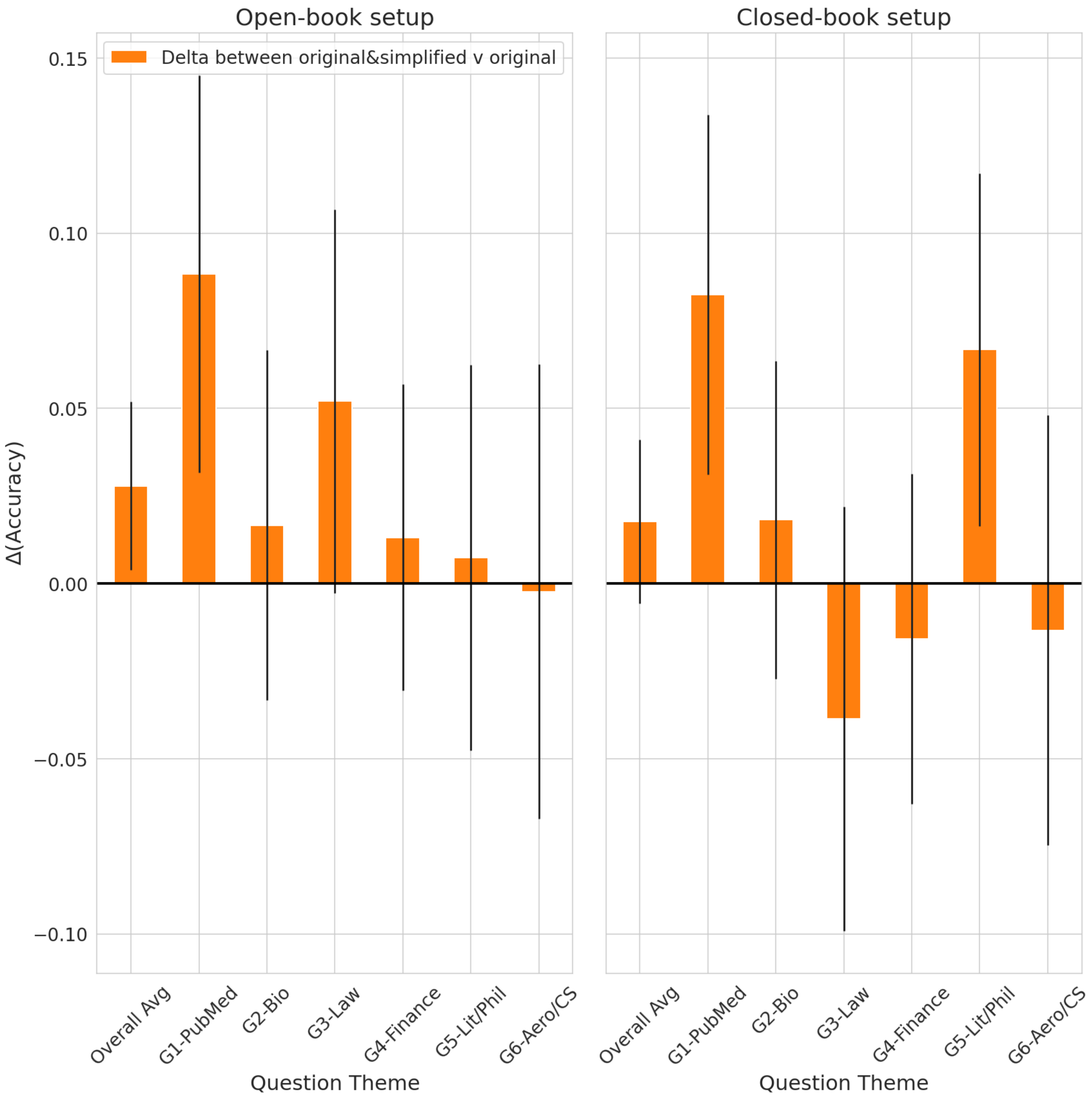

**Supplementary Figure 1: Change in multiple-choice question (MCQ) accuracy when participants viewed the original vs original+simplified texts.** This plot is the equivalent of Figure 3, but for arm 1 (original) vs 3 (original+simplified), and arm 4 (original) vs 6 (original+simplified).

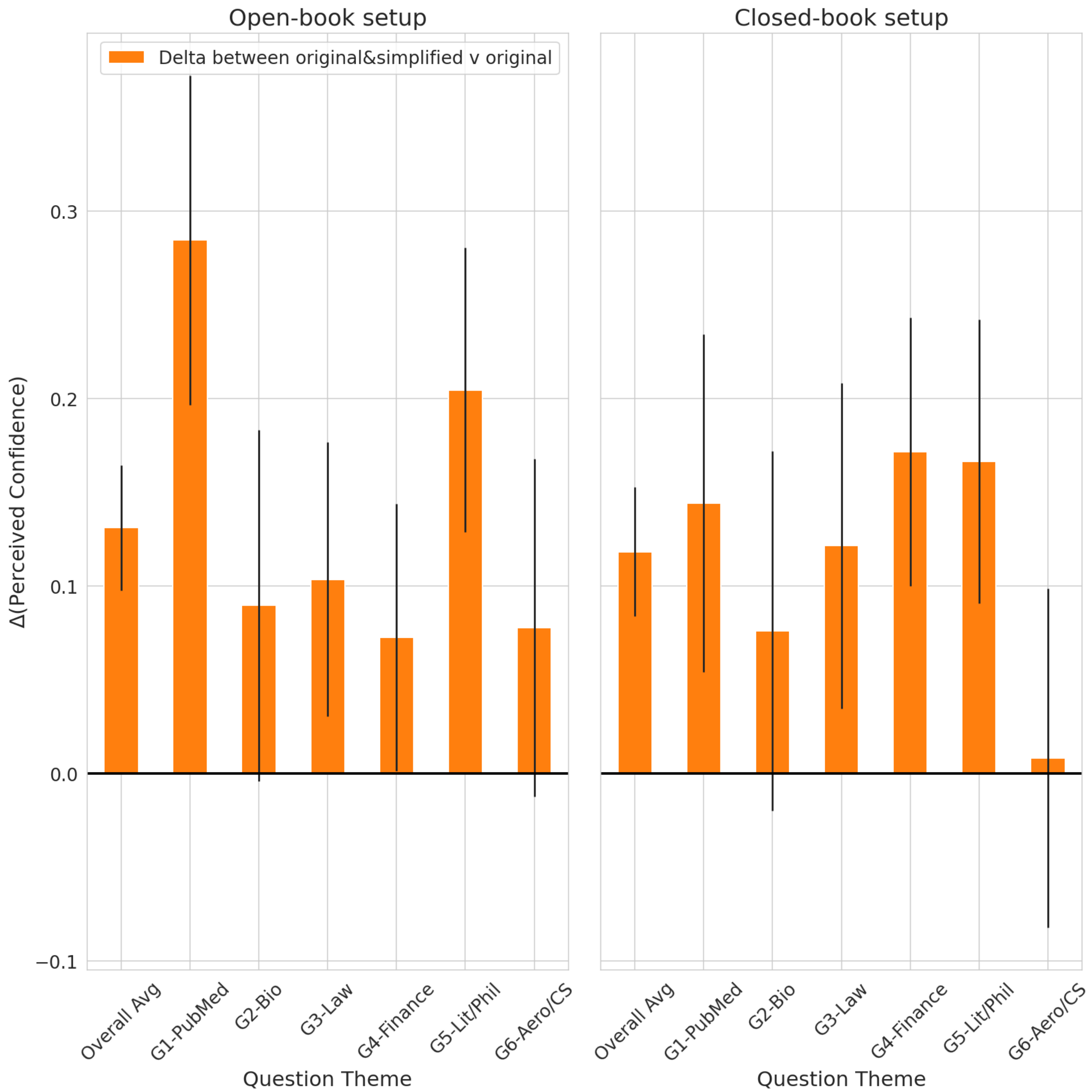

**Supplementary Figure 2: Change in perceived confidence per multiple-choice question (MCQ) when participants viewed the original vs original+simplified texts.** This plot is the equivalent of Figure 4, but for arm 1 (original) vs 3 (original+simplified), and arm 4 (original) vs 6 (original+simplified).

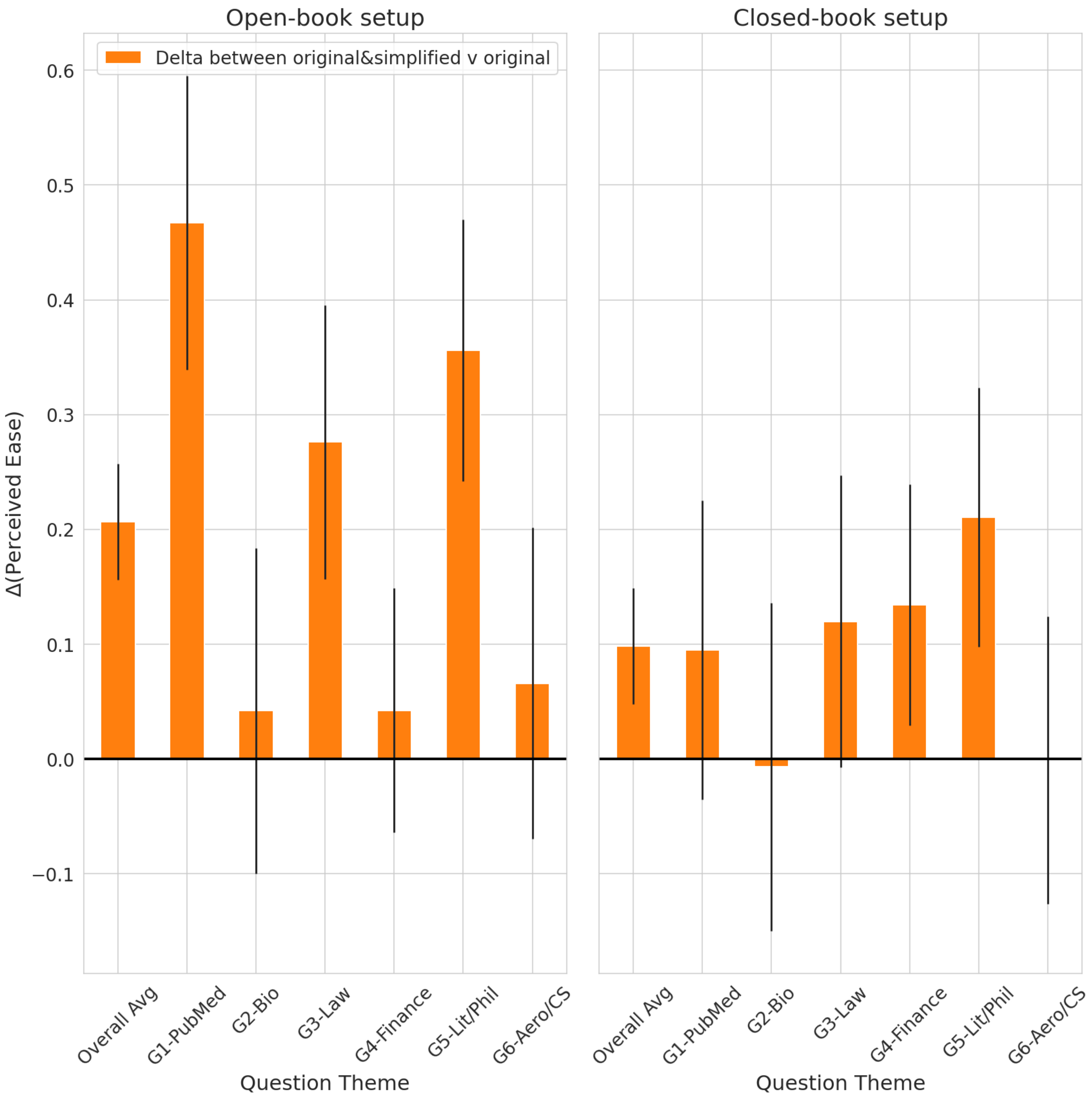

**Supplementary Figure 3: Change in perceived ease per text when participants viewed the original vs original+simplified texts.** This plot is the equivalent of Figure 5, but for arm 1 (original) vs 3 (original+simplified), and arm 4 (original) vs 6 (original+simplified).

Supplementary Data (all text excerpts and associated simplifications and MCQs)

Note: text excerpt in each row of <Original Text> is located at <Source URL> as of Dec 2, 2024, and licensed by their respective authors under the terms and conditions indicated therein. Text in <Simplified Text> columns are generated using our simplification model from <Original Text>. MCQ questions and answers were written by the authors of this paper, referring only to the <Original Text> field, and blinded to <Simplified Text>.

| Domain / topic | Source URL | Original Text | Simplified Text | MCQ 1 | Answer | MCQ 2 | Answer |
|---|---|---|---|---|---|---|---|
| PubMed | https://pmc.ncbi.nlm.nih.gov/articles/PMC11531917/ | Traditionally, blood slide microscopy has been the gold standard for malaria diagnosis. However, this method is time-consuming, requires well-trained personnel, and depends on adequate laboratory infrastructure, which is often lacking in endemic areas, particularly in resource-limited regions. Rapid diagnostic tests (RDTs) have been introduced as a more accessible alternative, offering faster results and easier application, especially in areas where laboratory capacity is limited.<br><br>RDTs work by detecting specific antigens produced by Plasmodium species in the blood. For example, histidine-rich protein 2 (HRP2) is specific to Pf, while aldolase and lactate dehydrogenase (LDH) can be detected in other Plasmodium species. This allows RDTs to differentiate between species, such as Pf-specific, Pv-specific, and pan-specific forms. However, there are still concerns regarding the sensitivity and specificity of RDTs, particularly in cases of low parasitemia or mixed infections. | Traditionally, looking at blood samples under a microscope has been the best way to find malaria. But this method takes a long time, needs people with special training, and depends on good labs, which are often missing in places where malaria is common, especially in areas with few resources. Rapid diagnostic tests, or RDTs, were introduced as a better option, giving results faster and being easier to use, especially where labs are not great.<br><br>RDTs work by finding specific proteins, called antigens, that the malaria parasite, Plasmodium, makes in the blood. For example, histidine-rich protein 2, or HRP2, is only found in Plasmodium falciparum (Pf), while aldolase and lactate dehydrogenase (LDH) can be found in other types of Plasmodium. This lets RDTs tell different types of malaria apart, like Pf-specific, Pv-specific, and those that detect any Plasmodium. However, there are still worries about how well RDTs find malaria (sensitivity) and how accurate they are (specificity), especially when there are very few parasites in the blood or when someone has more than one type of malaria. | What is the significance of HRP2 and LDH when it comes to rapid diagnostic tests (RDTs) for malaria?<br><br>a) They increase the accuracy of the tests by breaking down the malaria parasite.<br>b) They serve as identifying markers, enabling the tests to distinguish between different species of the malaria parasite.<br>c) They enhance the visibility of the test results, making it easier to interpret the outcome.<br>d) They are components of the human immune response that help fight off malaria infection.<br>e) They ensure the proper functioning of the test device by controlling the flow of fluids. | b | According to the text, which of the following is an area that rapid diagnostic tests (RDTs) for malaria have problems?<br><br>a) They can't tell if someone has more than one type of malaria parasite at the same time.<br>b) They need to be used in a special laboratory with lots of equipment.<br>c) They might not work as well when there aren't many malaria parasites in the blood.<br>d) They take longer to give results than looking at a blood sample under a microscope.<br>e) They can't tell you exactly which type of malaria someone has. | c |
| PubMed | https://pmc.ncbi.nlm.nih.gov/articles/PMC8036554/ | Checkpoint inhibitors such as PD-1 and PD-L1 demonstrate significant improvements in the clinic in some types of cancer, but their benefits are limited by fast post-therapy resistance and intrinsic absence of tumor neoantigens to enable blockade, rendering them ineffective for many patients. CAR T-cell therapy requires functional tumor-specific antigens on the cell surface of cancer cells to enable targeting, and lack of these challenge the therapeutic purpose of CAR T-cell therapy, especially in solid tumors, which often do not express such antigens and show a 9% overall response rate. | Checkpoint inhibitors like PD-1 and PD-L1, which help the body fight cancer, have shown improvements in some cancers. However, cancer cells can quickly become resistant to these treatments. Also, some cancers do not have the right proteins (neoantigens) for these inhibitors to work, making them ineffective for many patients. CAR T-cell therapy, a treatment that uses a patient's own immune cells to fight cancer, needs specific proteins (tumor-specific antigens) on the surface of cancer cells to work. Many solid tumors, a type of cancer that grows in organs, do not have these proteins. This limits the effectiveness of CAR T-cell therapy, with only a 9% success rate overall. | According to the passage, what's a challenge with using CAR T-cell therapy to treat cancer?<br><br>a) It only works for cancers that are resistant to other treatments.<br>b) Cancer can become resistant to this therapy after a while.<br>c) It needs special targets on cancer cells to work, and many cancers don't have them.<br>d) It's hard to give this therapy to patients.<br>e) It can make the cancer spread faster. | c | What do cancer cells need to have for checkpoint inhibitors to work effectively?<br><br>a) Specific tumor molecules that the drugs can recognize.<br>b) Large number of immune checkpoints.<br>c) CAR-T cells to assist.<br>d) A high level of genetic instability.<br>e) Limited ability to suppress the immune response. | a |
| PubMed | https://pmc.ncbi.nlm.nih.gov/articles/PMC10363436/ | In the intermittent access model, we examined the ability of semaglutide to decrease alcohol intake and block relapse-like drinking, as well as imaging the binding of fluorescently marked semaglutide to nucleus accumbens (NAc) in both male and female rats. The suppressive effect of semaglutide on alcohol-induced locomotor stimulation and in vivo dopamine release in NAc was tested in male mice. We evaluated effect of semaglutide on the in vivo release of dopamine metabolites (DOPAC and HVA) and gene expression of enzymes metabolising dopamine (MAOA and COMT) in male mice. | In the intermittent access model, we studied how semaglutide, a medicine, could reduce alcohol use and prevent a return to drinking in rats, both male and female. We also looked at how fluorescently marked semaglutide attached to the nucleus accumbens (NAc), a part of the brain involved in reward and motivation. In male mice, we tested how semaglutide affected movement caused by alcohol and dopamine release in the NAc. We also checked how semaglutide influenced the release of dopamine metabolites (DOPAC and HVA), chemicals that are created when dopamine is broken down, and the activity of genes that control enzymes (MAOA and COMT) involved in breaking down dopamine in male mice. | According to the passage, what is one way the researchers measured the effects of semaglutide on the male mice?<br><br>a) They observed how much alcohol the mice drank.<br>b) They measured how much the mice moved around after drinking alcohol.<br>c) They tracked the mice's sleeping patterns.<br>d) They measured the size of the mice's brains.<br>e) They observed how the mice interacted with each other. | b | According to the passage, what did the researchers do with semaglutide to help them see where it went in the rats' brains?<br><br>a) They gave the rats a very high dose of semaglutide.<br>b) They injected semaglutide directly into the rats' brains.<br>c) They added a special marker to semaglutide that glows.<br>d) They used a special type of semaglutide that only works in the brain.<br>e) They gave the rats semaglutide along with alcohol. | c |
| PubMed | https://pmc.ncbi.nlm.nih.gov/articles/PMC10177208/ | Combined pulmonary fibrosis and emphysema (CPFE) represents an increasingly recognized, progressive lung pathophysiology. The disease is associated with strong male predominance, a lack of specific treatment options and poor prognosis. The complex pathology of this condition involves emphysematous destruction of lung parenchyma, diffuse interstitial fibrosis, changes in the composition of lung immune cells, increased production of immunomodulatory factors and the prominent remodeling of pulmonary vasculature. Such an existence of obstructive and restrictive changes in the same lungs results in unexpectedly preserved lung volumes, while in contrast gas exchange is impaired. | Combined pulmonary fibrosis and emphysema (CPFE), a lung condition that's becoming more common, gets worse over time. It mostly affects men, has limited treatment options, and a poor outlook. This complex condition involves damage to the lung tissue from emphysema, a disease that damages the air sacs in the lungs, and widespread scarring of the lung tissue, called fibrosis. The immune cells in the lungs change, and the body makes more immunomodulatory factors, substances that control the immune system. The blood vessels in the lungs also change a lot. Since the lungs have both obstructive and restrictive problems, the lung volume stays about the same, which is unusual. However, the lungs cannot exchange gases properly. | According to the passage, what happens to the lungs in people with CPFE?<br><br>a) Lung tissue is destroyed, and scarring develops.<br>b) The lungs become smaller and less flexible.<br>c) The lungs become inflamed and filled with fluid.<br>d) Airways become blocked, and breathing becomes difficult.<br>e) The lungs produce excess mucus, leading to coughing. | a | According to the passage, what is the effect of having both obstructive and restrictive lung changes in CPFE?<br><br>a) It leads to a rapid decline in lung function.<br>b) It causes the lungs to become overinflated.<br>c) It results in unexpectedly normal lung size but poor oxygen exchange.<br>d) It makes it difficult to diagnose the condition accurately.<br>e) It increases the risk of developing lung cancer. | c |
| PubMed | https://pmc.ncbi.nlm.nih.gov/articles/PMC11475612/ | The liver presents a well-organized cellular architecture, which mirrors its complex and wide variety of functions. The liver is typified by the parenchymal cells (hepatocytes) and non-parenchymal cells. The latter include resident macrophages (Kupffer cells, KCs), hepatic stellate cells (HSC), lipocytes cells, and the sinusoidal intrahepatic lymphocytes (IHL). A signaling network connects parenchymal and non-parenchymal cells, | The liver has a very organized structure, which reflects its many complex jobs. The liver is made up of two main types of cells: parenchymal cells (hepatocytes, which are liver cells) and non-parenchymal cells. Non-parenchymal cells include Kupffer cells (KCs), which are a type of macrophage (a cell that eats up harmful things), hepatic stellate cells (HSC), lipocytes (fat-storing cells), and sinusoidal intrahepatic | According to the passage, what is one of the main causes of chronic liver disease?<br><br>a) A lack of communication between liver cells.<br>b) A decrease in the number of parenchymal cells.<br>c) An increase in the number of non-parenchymal | d | According to the passage, what can happen when the liver's normal metabolic processes are disrupted?<br><br>a) The liver can become enlarged and inflamed.<br>b) The liver can develop scar tissue and lose its function. | b |
| | | maintaining hepatic functions under homeostatic conditions. Alterations in hepatic and organ-related metabolic processes and pathways lead to changes in the hepatic histological spectrum, progressing toward chronic disease, which is accompanied by metabolically altered hepatocytes, inflammation, and fibrosis. | lymphocytes (IHL, a type of white blood cell in the liver). A network of signals connects the parenchymal and non-parenchymal cells, keeping the liver working properly under normal conditions. When the liver's and body's metabolic processes, or the ways the body uses energy, change, it can lead to changes in the liver's structure, which can lead to chronic liver disease. Chronic liver disease is accompanied by changes in liver cells, inflammation, and fibrosis (scarring). | cells.<br>d) Changes in the way the liver processes and uses energy.<br>e) A weakened immune response in the liver. | | c) The liver can become more efficient at filtering toxins.<br>d) The liver can regenerate damaged cells more quickly.<br>e) The liver can become less susceptible to diseases. | |
| Bio | https://www.gutenberg.org/cache/epub/1228/pg1228-images.html#chap01 | When we look to the individuals of the same variety or sub-variety of our older cultivated plants and animals, one of the first points which strikes us, is, that they generally differ much more from each other, than do the individuals of any one species or variety in a state of nature. When we reflect on the vast diversity of the plants and animals which have been cultivated, and which have varied during all ages under the most different climates and treatment, I think we are driven to conclude that this greater variability is simply due to our domestic productions having been raised under conditions of life not so uniform as, and somewhat different from, those to which the parent-species have been exposed under nature. There is, also, I think, some probability in the view propounded by Andrew Knight, that this variability may be partly connected with excess of food. It seems pretty clear that organic beings must be exposed during several generations to the new conditions of life to cause any appreciable amount of variation; and that when the organisation has once begun to vary, it generally continues to vary for many generations. | When we look at different plants or animals of the same type or subtype that we've grown for a long time, one of the first things we notice is that they differ more from each other than those found in the wild. When we think about the huge variety of plants and animals that have been cultivated and changed over time in different climates and with different care, I think we must conclude that this greater variety is simply because our domesticated plants and animals have grown in conditions that are not as consistent as, and are a bit different from, the conditions that their wild ancestors were used to. There's also a chance, as Andrew Knight suggested, that this variety might be partly related to having too much food. It seems pretty clear that living things need to be exposed to new living conditions for several generations to cause any noticeable change, and that once a living thing starts to change, it usually keeps changing for many generations. | According to the passage, what is the main difference between plants and animals raised by humans and those in the wild?<br><br>a) Ones raised by humans tend to be more diverse.<br>b) Ones raised by humans are less adaptable to changing conditions.<br>c) Ones raised by humans have a shorter lifespan.<br>d) Ones raised by humans are more resistant to disease.<br>e) Ones raised by humans are generally smaller in size. | a | According to the text, what might explain the greater variability in plants and animals raised by humans?<br><br>a) Differences in environmental conditions.<br>b) Genetic mutations caused by artificial selection.<br>c) Crossbreeding with different species.<br>d) Exposure to pesticides and herbicides.<br>e) Lack of competition for resources. | a |
| Bio | https://www.genome.gov/27528684/1000-genomes-project | The 1000 Genomes Project is a collaboration among research groups in the US, UK, and China and Germany to produce an extensive catalog of human genetic variation that will support future medical research studies. It will extend the data from the International HapMap Project, which created a resource that has been used to find more than 100 regions of the genome that are associated with common human diseases such as coronary artery disease and diabetes. The goal of the 1000 Genomes Project is to provide a resource of almost all variants, including SNPs and structural variants, and their haplotype contexts. This resource will allow genome-wide association studies to focus on almost all variants that exist in regions found to be associated with disease. The genomes of over 1000 unidentified individuals from around the world will be sequenced using next generation sequencing technologies. The results of the study will be publicly accessible to researchers worldwide. | The 1000 Genomes Project is a partnership between research groups in the United States, the United Kingdom, China, and Germany. The goal is to create a huge list of human genetic differences, or variations, that will help future medical research. It will build on the International HapMap Project, which created a tool used to find over 100 areas in our genes linked to common diseases like coronary artery disease, a heart condition, and diabetes, a disease where the body doesn't control blood sugar well.<br><br>The 1000 Genomes Project wants to make a resource that includes almost all gene variations, including SNPs (single nucleotide polymorphisms, small changes in DNA) and structural variants (larger changes in DNA), and how they are connected. This resource will help researchers study the whole genome and focus on nearly all the variations found in areas linked to diseases. The genes of over 1000 people from different parts of the world, whose identities are not revealed, will be studied using new sequencing technologies. The results of the study will be available to researchers all over the world. | The goal of the 1000 Genomes Project is to:<br><br>a) Find all genome regions related to disease.<br>b) Find 1000 genome regions related to disease.<br>c) Find all genomic variants or changes related to common human diseases.<br>d) Find all genomic variants or changes related to disease.<br>e) Find all genomic variants or changes. | e | What is the relationship of the 1000 Genomes Project to the International HapMap Project?<br><br>a) 1000 Genomes extends HapMap data from 100 to 1000 genomes.<br>b) 1000 Genomes extends the types of genetic variants and genomic regions covered in HapMap.<br>c) 1000 Genomes examines genetic variants related to coronary artery disease and diabetes in more detail.<br>d) 1000 Genomes includes more countries than HapMap.<br>e) All of the above | b |
| Bio | https://en.wikinews.org/wiki/2006_Nobel_Prize_in_Medicine_awarded_to_American_scientists | The statement from the Nobel Assembly of the Karolinska Institute in Sweden said: "This year's Nobel Laureates have discovered a fundamental mechanism for controlling the flow of genetic information." Fire and Mello's (then at Washington's Carnegie Institution) seminal publication in Nature in 1998 opened the door for "exciting possibilities", the jury in Stockholm added.<br><br>RNA interference (or RNAi) is the process of using double stranded RNA fragments which bind and interfere with a specific messenger RNA, so that it's not longer used to make proteins. It has been recognised as a natural way of gene regulation in plants and animals. Today, the technology is being used by biomedical scientists to fiddle genes involved in diseases such as cancer, and to prevent infection with hepatitis viruses. | The Nobel Assembly at the Karolinska Institute in Sweden said, "This year's winners found a key way to control how genetic information moves." Fire and Mello, who were at the Carnegie Institution in Washington at the time, wrote a very important paper in the science journal Nature in 1998. This paper opened the door to many exciting possibilities, the judges in Stockholm said.<br><br>RNA interference, also known as RNAi, is a process where small pieces of double-stranded RNA attach to a specific messenger RNA (mRNA). Messenger RNA is a molecule that carries instructions for making proteins. When the RNAi attaches to the mRNA, it stops the mRNA from being used to make proteins. Scientists have learned that this is a natural way for plants and animals to control their genes. Now, biomedical scientists are using this technology to change genes involved in diseases like cancer and to stop hepatitis viruses from causing infections. | What was the Nobel prize in this passage awarded for?<br><br>a) The Nature publication in 1998.<br>b) Discovering a new way to control genetic information.<br>c) Discovering how proteins are made in plants and animals.<br>d) Discovering how genetic information flows between generations of plants and animals.<br>e) Modifying genes for cancer and hepatitis. | b | What is RNA interference?<br><br>a) How RNA can interact to stop protein production from RNA.<br>b) How plants and animals use protein to stop RNA from forming.<br>c) How scientists cure diseases such as cancer and hepatitis.<br>d) How RNA become protein.<br>e) How viruses and cancer interfere with the formation of protein using RNA. | a |
| Bio | https://en.wikinews.org/wiki/%27Earth-based_life_can_survive_in_hydrogen-rich_atmospheres%27:_MIT_professor_Dr_Seager_tells_Wikinews_about_her_research_on_organisms_thriving_in_oxygen-less_environmen | The study reported the organisms [Escherichia coli strain K-12 and Saccharomyces cerevisiae strain S288C] were reproducing normally in both 100% H2 and 100% He environment. However, the sigmoid-shaped growth curve was not on par with 100% air. E. coli and yeast switch from aerobic respiration, which uses oxygen, to anaerobic respiration and fermentation. Both processes are less efficient and do not produce as much energy as aerobic respiration.<br><br>E. coli in an 80%-20% N2-CO2 environment had slower growth rate as CO2 dissolves and makes the liquid medium acidic. Such reduction in growth rate was not observed for yeast cultures, which can thrive in acidic environments. However, yeast's growth rate in 100% air was far greater than in the other three media. The likely reason for this significant difference was lack of oxygen for non-respiratory purposes, the research reported. Oxygen is essential for synthesis of biochemicals | The study found that Escherichia coli (a type of bacteria) and Saccharomyces cerevisiae (a type of yeast) reproduced normally in environments with only hydrogen gas (H2) or helium gas (He). However, their growth wasn't as good as it was in air. E. coli and yeast usually use oxygen for energy through a process called aerobic respiration. When oxygen is not available, they switch to anaerobic respiration and fermentation, which are less efficient and produce less energy.<br><br>In an environment with 80% nitrogen and 20% carbon dioxide, E. coli grew more slowly. This is because the carbon dioxide dissolved in the liquid, making it acidic. Yeast, however, didn't show this slow-down because it can grow in acidic conditions. But, yeast grew much faster in air than in the other three environments. The study suggests that this is because oxygen is needed for things other than respiration. Oxygen is important for making chemicals like heme and sterols, which are | What is true about the organisms growing in pure H2 or He environments, compared to growing in air?<br><br>a) They grew faster, with a sigmoid curve.<br>b) They stopped using oxygen.<br>c) They started aerobic respiration.<br>d) They learned to use acid to generate energy.<br>e) They used stored oxygen to survive. | b | Why did the growth rate change for the some organisms in the N2-CO2 environment?<br><br>a) Some organisms did not do well with acid.<br>b) Some organisms did well in acid.<br>c) Some organisms made too much heme and sterols.<br>d) Some organisms stored oxygen in their synthesized products and continued to grow, while others could not.<br>e) All of the above. | a |

| | | | | | | | | |
|---|---|---|---|---|---|---|---|---|
| | t | such as heme and sterols, which are important for yeast. In atmospheres lacking oxygen to produce these chemicals, yeast fungi have stunted growth rate. | important for yeast. When there's not enough oxygen to make these chemicals, yeast growth slows down. | | | | |
| Law | https://en.wikipedia.org/wiki/Specific_performance | An injunction, often concerning confidential information or real property, is a type or subset of specific performance and is one of the more commonly-used forms of specific performance. While specific performance can be in the form of any type of forced action, it is usually to complete a previously established transaction, thus being the most effective remedy in protecting the expectation interest of the innocent party to a contract. It is usually the opposite of a prohibitory injunction, but there are mandatory injunctions that have a similar effect to specific performance and these kinds of distinctions are often difficult to apply in practice or even illusory. | An injunction, a court order often related to secret information or land, is a kind of specific performance. It's one of the most common types of specific performance. Specific performance can be any type of forced action, but it's usually used to finish a deal that was already agreed upon. This makes it the best way to protect the innocent party's expectation of what they would get from the contract. It's usually the opposite of a prohibitory injunction, which stops someone from doing something. However, there are mandatory injunctions that have a similar effect to specific performance. These differences are often hard to apply in real situations or may not even be real differences. | What is the main purpose of "specific performance," as described in the passage?<br><br>a) To punish someone who has broken a contract.<br>b) To force someone to complete a transaction they agreed to.<br>c) To prevent someone from speaking negatively about their company.<br>d) To resolve a dispute over paying back a loan.<br>e) To protect the financial interests of both parties in a contract. | b | According to the passage, what is a common example of when an injunction might be used?<br><br>a) When someone breaks a promise to deliver goods on time.<br>b) When someone tries to sell a piece of land that they don't own.<br>c) When someone reveals private information about a company.<br>d) When someone refuses to pay a bill that they owe.<br>e) When someone damages another person's property. | c |
| Law | https://en.wikipedia.org/wiki/Parol_evidence_rule | The parol evidence rule is a rule in common law jurisdictions limiting the kinds of evidence parties to a contract dispute can introduce when trying to determine the specific terms of a contract and precluding parties who have reduced their agreement to a final written document from later introducing other evidence, such as the content of oral discussions from earlier in the negotiation process, as evidence of a different intent as to the terms of the contract. The rule provides that "extrinsic evidence is inadmissible to vary a written contract". The term "parol" derives from the Anglo-Norman French parol or parole, meaning "word of mouth" or "verbal", and in medieval times referred to oral pleadings in a court case. | The parol evidence rule, a rule used in legal systems based on common law, limits the types of evidence that people involved in a contract dispute can use. It's about figuring out the exact terms of a contract. This rule stops people who have a written agreement from bringing in other evidence, like conversations they had earlier during negotiations, to change the meaning of the contract. The rule says that evidence from outside the written contract cannot be used to change it. The word "parol" comes from Anglo-Norman French, meaning "word of mouth" or "spoken." In the past, it referred to spoken arguments in court cases. | What is the "parol evidence rule," according to the passage?<br><br>a) A rule that prevents people from lying in court.<br>b) A rule that limits what evidence can be used in contract disputes.<br>c) A rule that forces people to follow through on their promises.<br>d) A rule that helps people negotiate fair agreements.<br>e) A rule that allows people to change their minds about a contract. | b | According to the passage, what does the parol evidence rule prevent people from doing?<br><br>a) Using written notes as evidence in court.<br>b) Introducing evidence that contradicts a written contract.<br>c) Talking about their agreements with other people.<br>d) Changing the terms of a contract after it's been signed.<br>e) Getting legal help with a contract dispute. | b |
| Law | https://en.wikipedia.org/wiki/Estoppel | Estoppel is sometimes said to be a rule of evidence whereby a person is barred from leading evidence of a fact that has already been settled or they are otherwise precluded from asserting, but that may be an oversimplification. Firstly, although some estoppels relate to preventing a party from asserting facts, others relate to preventing a party from asserting a right or a claim. Secondly, under the conflict of laws in common law jurisdictions, matters of evidence are usually treated as procedural matters for the law of the local court (the lex fori), whereas it is generally accepted that an estoppel may affect substantive rights and are therefore matters to be determined by the proper law (or lex causae) that governs the particular issue. | Estoppel, a legal principle, is sometimes described as a rule of evidence that stops someone from presenting proof of something already decided or prevents them from making certain claims. However, this might be too simple. First, while some estoppels stop people from claiming facts, others stop them from claiming rights or making demands. Second, in common law countries, evidence rules are usually considered procedural matters that follow the local court's laws (the lex fori). But generally, an estoppel can affect important rights, and these are usually decided by the proper law (lex causae) that relates to the specific issue. | According to the passage, what is estoppel?<br><br>a) A rule that prevents someone from saying something that was already decided.<br>b) A rule that forces someone to do something they promised to do.<br>c) A rule that punishes someone for breaking a contract.<br>d) A rule that helps people negotiate a fair agreement.<br>e) A rule that allows someone to change their mind about a promise. | a | According to the passage, when a court decides whether to use the rule of estoppel, which of these matter most?<br><br>a) The laws of the place where the court is located.<br>b) The laws that specifically deal with the issue in the case.<br>c) What the judge thinks is the right way to decide.<br>d) How much evidence there is to prove what happened.<br>e) The roles that each person involved in the case played. | b |
| Law | https://www.consumer.gov/sites/www.consumer.gov/files/articles/pdf/pdf-1050g-sample_rental_agreement_basic_beginning_renting_an_apartment_or_house.pdf | The Tenants signing this Rental Contract hereby state that all questions about this Rental Agreement have been answered, that they fully understand all the provisions of the agreement and the obligations and responsibilities of each party, as spelled out herein. They further state that they agree to fulfill their obligations in every respect or suffer the full legal and financial consequences of their actions or lack of action in violation of this agreement. Signature by the Tenant on this Rental Agreement is acknowledgement and he/she has received a signed copy of the Rental Agreement. | The tenants signing this Rental Contract say that all their questions about it have been answered. They also say they understand all the parts of the agreement and what each person involved is responsible for, as explained in the contract. They agree to do what they are supposed to do. If they don't, they will face legal and financial problems because they broke the agreement. When a tenant signs this Rental Contract, it means they have gotten a signed copy of it. | According to the rental contract, what happens if the tenants don't follow the rules?<br><br>a) They might have to pay extra money or face other consequences.<br>b) The landlord will remind them of their obligations.<br>c) They can talk to the landlord to change the rules.<br>d) Nothing will happen as long as they pay their rent on time.<br>e) They can move out without giving any notice. | a | What does the rental contract say about the tenants' responsibilities?<br><br>a) The tenants are responsible for understanding the agreement.<br>b) The landlord is responsible for explaining the agreement to the tenants.<br>c) The tenants can ask a lawyer to explain their responsibilities.<br>d) The tenants don't have any responsibilities until they move in.<br>e) The tenants can decide which responsibilities they want to follow. | a |
| Law | https://www.dre.ca.gov/files/pdf/re6.pdf | This garage door opener or child resistant pool barrier may not be in compliance with the safety standards relating to automatic reversing devices as set forth in Chapter 12.5 (commencing with Section 19890) of Part 3 of Division 13 of, or with the pool safety standards of Article 2.5 (commencing with Section 115920) of Chapter 5 of Part 10 of Division 104 of, the Health and Safety Code. The water heater may not be anchored, braced, or strapped in accordance with Section 19211 of the Health and Safety Code. Window security bars may not have quick-release mechanisms in compliance with the 1995 Edition of the California Building Standards Code. | This garage door opener or child-resistant pool barrier might not meet safety standards for automatic reversing devices, as explained in Chapter 12.5 (starting with Section 19890) of the Health and Safety Code. It also might not meet pool safety standards in Article 2.5 (starting with Section 115920) of the same code. The water heater might not be fastened properly, as described in Section 19211 of the Health and Safety Code. Window security bars might not have quick-release mechanisms that meet the 1995 California Building Standards Code. | According to the passage, why might window security bars not be safe?<br><br>a) They might be made of weak or flimsy materials.<br>b) They might not be properly attached to the window frame.<br>c) They might not have a way to open them quickly in an emergency.<br>d) They might prevent people from entering the home in a fire.<br>e) They might be too difficult for children to open. | c | According to the passage, why might the pool not be safe?<br><br>a) It might not be deep enough<br>b) The water might not be clean.<br>c) The barrier might automatically close.<br>d) There might not be a lifeguard on duty.<br>e) It might be too cold to swim. | c |
| Finance | https://en.wikipedia.org/wiki/Amortization_(accounting) | In accounting, amortization is a method of obtaining the expenses incurred by an intangible asset arising from a decline in value as a result of use or the passage of time. Amortization is the acquisition cost minus the residual value of an asset, calculated in a systematic manner over an asset's useful economic life. Depreciation is a corresponding concept for tangible assets.<br><br>While theoretically amortization is used to account for the decreasing value of an intangible asset over its useful life, in practice many companies will amortize what would otherwise be one-time expenses | In accounting, amortization is a way to record the costs of an intangible asset, like a patent or a trademark, as it loses value over time or due to use. Amortization is figured out by subtracting the asset's remaining value from its original cost and then spreading that cost out over the asset's useful life, which is how long it's expected to be helpful to the business. Depreciation is a similar idea, but it's used for tangible assets, things you can touch, like buildings or machines.<br><br>Although the idea of amortization is to show how an intangible asset's value goes down over time, many businesses use it in a different way. | What is the difference between amortization and depreciation?<br><br>a) Amortization is for things you can't touch, like patents, and depreciation is for things you can, like a building.<br>b) Amortization is used to lower a company's taxes, while depreciation is used to calculate the company's profits.<br>c) Amortization is for things that lose value quickly, | a | According to the passage, how do some companies use amortization?<br><br>a) To make the value of things like patents and trademarks go down over time.<br>b) To spread out the cost of one-time expenses and make their profits look better.<br>c) To make the value of things like buildings and equipment go up.<br>d) To avoid paying taxes on their expenses. | b |
| | | through listing them as a capital expense on the cash flow statement and paying off the cost through amortization, having the effect of improving the company's net income in the fiscal year or quarter of the expense. | They take one-time expenses and spread them out over time by listing them as a capital expense on the cash flow statement and paying them off through amortization. This makes the company's net income, or profit, look better for that year or quarter when the expense was first recorded. | while depreciation is for things that last a long time.<br>d) Amortization is when a company buys something, and depreciation is when a company sells something.<br>e) They're basically the same thing; there's no real difference. | | e) To make their financial reports more accurate. | |
| Finance | https://en.wikipedia.org/wiki/Modern_portfolio_theory | Modern portfolio theory (MPT), or mean-variance analysis, is a mathematical framework for assembling a portfolio of assets such that the expected return is maximized for a given level of risk. It is a formalization and extension of diversification in investing, the idea that owning different kinds of financial assets is less risky than owning only one type. Its key insight is that an asset's risk and return should not be assessed by itself, but by how it contributes to a portfolio's overall risk and return. The variance of return (or its transformation, the standard deviation) is used as a measure of risk, because it is tractable when assets are combined into portfolios. Often, the historical variance and covariance of returns is used as a proxy for the forward-looking versions of these quantities, but other, more sophisticated methods are available. | Modern portfolio theory (MPT), also known as mean-variance analysis, is a mathematical way to put together a group of investments, called a portfolio. The goal is to get the highest possible return for a certain amount of risk. It's like a more advanced and detailed version of diversification in investing, which is the idea that having different types of investments is safer than just having one type. The main idea of MPT is that you shouldn't look at an investment's risk and return by itself. Instead, you should consider how it affects the whole portfolio's risk and return. The variance of return, or its related measure, the standard deviation, is used to measure risk because it's easy to work with when you combine different investments into a portfolio. Usually, the past variance and covariance of returns are used to guess what the future risk will be. However, there are other, more complex methods that can be used. | According to the passage, what is the main idea of Modern Portfolio Theory?<br><br>a) Put all your money in one thing that could earn a lot.<br>b) Try to earn the most money by taking big risks.<br>c) Spread your money around so you won't lose any of it.<br>d) Guess what the market will do and buy and sell at the right time.<br>e) Find the best mix of investments to earn more with less risk. | e | According to the passage, how does Modern Portfolio Theory determine the risk of an investment?<br><br>a) By measuring how much its value goes up and down.<br>b) By considering the size of the company.<br>c) By analyzing the number of investors.<br>d) By predicting future market performance.<br>e) By evaluating expert opinions on the investment. | a |
| Finance | https://en.wikipedia.org/wiki/Dividend | A dividend is a distribution of profits by a corporation to its shareholders, after which the stock exchange decreases the price of the stock by the dividend to remove volatility. The market has no control over the stock price on open on the ex-dividend date, though more often than not it may open higher. When a corporation earns a profit or surplus, it is able to pay a portion of the profit as a dividend to shareholders. Any amount not distributed is taken to be re-invested in the business (called retained earnings). The current year profit as well as the retained earnings of previous years are available for distribution; a corporation is usually prohibited from paying a dividend out of its capital. | A dividend is when a company shares its profits with the people who own parts of the company, called shareholders. After the company pays the dividend, the stock exchange, a place where stocks are bought and sold, lowers the stock price by the amount of the dividend. This is done to reduce big price swings, also known as volatility. On the ex-dividend date, the day the stock starts trading without the right to the dividend, the market doesn't control the stock's opening price. However, it often opens higher.<br><br>When a company makes a profit, it can share some of that money as a dividend with its shareholders. Any profit that's not given out as a dividend is kept by the company and used for future growth, called retained earnings. The company can use its current year's profit and the profits it kept from past years to pay dividends. Usually, a company is not allowed to pay a dividend using its main funds, also known as its capital. | According to the passage, what can a company do with its profits?<br><br>a) Donate them to charity.<br>b) Pay them to shareholders as stock.<br>c) Use them to buy back its own stock.<br>d) Reinvest them in the business or pay them as dividends.<br>e) Use them to pay off all its debts. | d | According to the passage, which of these is a source of money that a company can use to pay dividends to its shareholders?<br><br>a) Money the company made this year, before paying its bills.<br>b) Money the company made in past years that it didn't pay to shareholders.<br>c) Money the company gets from selling its stock on the stock market.<br>d) Money the company has borrowed but hasn't paid back yet.<br>e) The value of everything the company owns, like its buildings and equipment. | b |
| Finance | https://www.irs.gov/taxtopics/tc501 | You cannot take the standard deduction if:<br><br>* You are a married individual filing as married filing separately whose spouse itemizes deductions.<br>* You are an individual who files a tax return for a period of less than 12 months because of a change in your annual accounting period.<br>* You were a nonresident alien or a dual-status alien during the year. However, nonresident aliens who are married to a U.S. citizen or resident alien at the end of the year and who choose to be treated as U.S. residents for tax purposes can take the standard deduction. For additional information, refer to Publication 519, U.S. Tax Guide for Aliens.<br>* You are filing as an estate or trust, common trust fund, or partnership.<br><br>You should itemize deductions on Schedule A (Form 1040), Itemized Deductions if the total amount of your allowable itemized deductions is greater than your standard deduction or if you must itemize deductions because you can't use the standard deduction. You may also want to itemize deductions if your standard deduction is limited because another taxpayer claims you as a dependent. Itemized deductions, subject to certain dollar limitations, include amounts you paid, during the taxable year, for state and local income or sales taxes, real property taxes, personal property taxes, mortgage interest, disaster losses, gifts to charities, and medical and dental expenses. | You cannot use the standard deduction if:<br><br>* You are married and file separately, and your spouse uses itemized deductions.<br>* You file a tax return for less than a full year (12 months) because you changed your accounting period.<br>* You were a nonresident alien (someone not a US citizen living in the US) or a dual-status alien (someone who was both a resident and nonresident) during the year. However, if a nonresident alien is married to a US citizen or resident at the end of the year and chooses to be treated as a US resident for taxes, they can use the standard deduction. For more details, see Publication 519, U.S. Tax Guide for Aliens.<br>* You are filing as an estate, trust, common trust fund, or partnership.<br><br><br>You should use Schedule A (Form 1040), Itemized Deductions, if your total allowable itemized deductions are more than your standard deduction or if you cannot use the standard deduction. You might also want to use itemized deductions if your standard deduction is limited because someone else claims you as a dependent. Itemized deductions, up to certain amounts, include things you paid during the tax year, like state and local income or sales taxes, property taxes, mortgage interest, disaster losses, charitable donations, and medical and dental expenses. | The passage suggests that someone might choose to itemize deductions if:<br><br>a) They want to avoid calculating their standard deduction.<br>b) They are unsure whether they qualify for the standard deduction.<br>c) They want to maximize their tax refund.<br>d) They have a simple tax situation with few deductions.<br>e) They are filing their taxes for the first time. | c | According to the passage, what might limit someone's standard deduction?<br><br>a) Lots of expenses you can subtract from your income.<br>b) Someone else puts you on their taxes as someone they support financially.<br>c) You didn't earn any money that you have to pay taxes on.<br>d) Sending in your taxes after the deadline.<br>e) Living in a state where you have to pay a lot of income tax. | b |
| Finance | https://en.wikipedia.org/wiki/Capital_gains_tax_in_the_United_States#:~:text=Capital%20gains%20taxes%20are%20disproportionately,that%20generate%20the%20taxable%20gains. | Capital gains taxes are disproportionately paid by high-income households, since they are more likely to own assets that generate the taxable gains. While this supports the argument that payers of capital gains taxes have more "ability to pay", it also means that the payers are especially able to defer or avoid the tax, as it only comes due if and when the owner sells the asset.<br><br>Low-income taxpayers who do not pay capital gains taxes directly may wind up paying them through changed prices as the actual payers pass through the cost of paying the tax. Another factor complicating the use of capital gains taxes to address income inequality is that capital gains are usually not recurring income. A taxpayer may be "high-income" in the single year in which he or she sells an asset or invention. | People with high incomes pay a larger share of capital gains taxes because they're more likely to own things like stocks or property that can increase in value and be taxed. This supports the idea that people who pay capital gains taxes can afford to pay more, but it also means they can easily delay or avoid the tax. They only pay when they sell the asset.<br><br><br>People with low incomes who don't pay capital gains taxes might end up paying them indirectly through higher prices. This happens when people who do pay the tax pass the cost on to others. Another thing that makes it hard to use capital gains taxes to fix income inequality is that capital gains aren't usually a regular income. Someone might have a high income in a single year if they sell a valuable item or invention. | What is one of the reasons why capital gains taxes may not effectively address income inequality?<br><br>a) They are not progressive, meaning that everyone pays the same rate.<br>b) They can be easily avoided by high-income households.<br>c) They disproportionately impact low-income households.<br>d) They discourage investment and economic growth.<br>e) They are too complex to administer effectively. | b | According to the text, how can capital gains taxes indirectly impact low-income taxpayers?<br><br>a) By increasing the cost of goods and services.<br>b) By reducing the value of their investments.<br>c) By making it harder for them to qualify for government assistance.<br>d) By increasing their tax burden on other forms of income.<br>e) By discouraging businesses from hiring new employees. | a |
| Finance | https://en.wikinews.org/wiki/US_u | The United States has now lost 7.2 million jobs since the recession officially began in December 2007. The new data has sparked fears that | The United States has lost 7.2 million jobs since the recession, which officially started in December 2007. The new job numbers have caused | Which aspects are NOT barriers towards improving financial conditions? | b | What warnings are NOT provided for economic recovery above? | d |
| | nemployment_rate_reaches_9.8%25 | unemployment could threaten an economic recovery. Top US officials have warned that any recovery would be slow and uneven, and some have predicted the unemployment rate will top 10% before the situation improves.<br><br>"Continued household deleveraging and rising unemployment may weigh more on consumption than forecast, and accelerating corporate and commercial property defaults could slow the improvement in financial conditions," read a report by the International Monetary Fund's World Economic Outlook, predicting that unemployment will average 10.1% by next year and not go back down to five percent until 2014.<br><br>Mark Zandi, chief economist at Moody's Economy.com, said that "it's a very fragile and tentative recovery. Policy makers need to do more." | worry that unemployment could stop the economy from getting better. Top US officials have said that any economic recovery will be slow and uneven. Some have even said that the unemployment rate, the percentage of people without jobs, will go above 10% before things get better.<br><br>A report from the International Monetary Fund (IMF), a group that helps countries with economic issues, said that people paying off debt and rising unemployment could hurt spending more than expected. The IMF also said that more businesses and property owners failing to pay their loans could slow down improvements in the economy. The report predicts that the average unemployment rate will be 10.1% next year and won't go back down to 5% until 2014.<br><br>Mark Zandi, the chief economist at Moody's Economy.com, a company that analyzes the economy, said that the recovery is weak and uncertain. He also said that government officials need to do more to help the economy. | a) Households reducing debt.<br>b) Households borrowing more money.<br>c) More people unable to find jobs.<br>d) Companies not paying their real estate rent.<br>e) Companies not paying their real estate mortgages. | | a) Recovery may be different in different places.<br>b) Recovery may be slow.<br>c) More than 1 in 10 people may be unable to find jobs<br>d) People may not be able to pay off their debts.<br>e) Recovery may not happen. | |
| Finance | https://en.wikinews.org/wiki/UK_energy_companies_announce_that_prices_for_bills_could_increase | The "big six" energy companies in the United Kingdom are British Gas, E-on, Npower, Scottish and Southern Energy, Scottish Power, and EDF Energy. British Gas stated: "Prices [are] likely to remain at historically high levels, and in fact likely to increase as non-commodity costs rise ever upwards."<br><br>EDF Energy said: "[We] would of course be prepared to reduce tariffs if market conditions allowed." Scottish Power stated: "There are no immediate signals that would indicate a fall in retail prices for this winter, and risks of an increase next year." Scottish & Southern Energy commented: "With forward annual wholesale prices significantly higher, and with upward pressures in terms of distribution, environmental and social costs, seeking to avoid an increase between now and the end of 2010 is an important goal."<br><br>Meanwhile, a study by Consumer Focus in early September 2009 suggested that "energy companies were overcharging customers by £100 ($162) every year. A spokesperson for Ofgem said that there was no evidence of any cartel in operation, or evidence of profiteering. The spokesperson commented: "It is up to the companies themselves to decide whether to cut their bills. Consumer Focus data suggests that Scottish Power has increased dual fuel prices by the most since 2003 - up 148% - while decreasing prices by 0.6% so far this year. RWE's Npower has increased tariffs by 132% since 2003, but has reduced bills by 2.7% in 2009." | The six biggest energy companies in the United Kingdom, also known as the "big six," are British Gas, E-on, Npower, Scottish and Southern Energy, Scottish Power, and EDF Energy. British Gas said that energy prices will probably stay very high and might even go up because of rising costs that are not related to the price of energy itself.<br><br>EDF Energy said that they would lower their prices if the market allowed them to. Scottish Power said that there are no signs that energy prices will fall this winter, and there is a chance they could rise next year. Scottish and Southern Energy said that because the future price of energy is much higher, and because of rising costs for things like getting energy to people, environmental issues, and social issues, they want to avoid raising prices until the end of 2010.<br><br>Meanwhile, a study done in early September 2009 by Consumer Focus, a consumer protection organization, suggested that energy companies were charging customers too much, about 100 pounds or 162 dollars every year. A spokesperson for Ofgem, a government agency that regulates the energy market, said there was no proof that the companies were working together to set prices or that they were making too much money. The spokesperson said that it was up to the companies to decide whether to lower their bills. Consumer Focus data showed that Scottish Power has raised prices for gas and electricity the most since 2003, by 148 percent, but has only lowered prices by 0.6 percent this year. RWE's Npower has raised prices by 132 percent since 2003, but has lowered them by 2.7 percent in 2009. | Why do we need to try and avoid future price increases for energy bills?<br><br>a) Because it is already high.<br>b) Because the cost of generating energy is expected to increase.<br>c) Because people don't want to pay more.<br>d) Because the energy companies are getting fined more.<br>e) Because companies are working together to increase prices. | b | When will energy bills fall?<br><br>a) When costs of energy production drops.<br>b) When the companies stop working together to maintain high prices.<br>c) When environmental fines decrease.<br>d) When people demand it.<br>e) When the companies decide to lower prices. | e |
| Lit/Phil | https://www.gutenberg.org/cache/epub/174/pg174-images.html#chap00 | From the corner of the divan of Persian saddle-bags on which he was lying, smoking, as was his custom, innumerable cigarettes, Lord Henry Wotton could just catch the gleam of the honey-sweet and honey-coloured blossoms of a laburnum, whose tremulous branches seemed hardly able to bear the burden of a beauty so flamelike as theirs; and now and then the fantastic shadows of birds in flight flitted across the long tussore-silk curtains that were stretched in front of the huge window, producing a kind of momentary Japanese effect, and making him think of those pallid, jade-faced painters of Tokyo who, through the medium of an art that is necessarily immobile, seek to convey the sense of swiftness and motion. The sullen murmur of the bees shouldering their way through the long unmown grass, or circling with monotonous insistence round the dusty gilt horns of the straggling woodbine, seemed to make the stillness more oppressive. | From the corner of the long, cushioned seat (a divan) made of Persian saddlebags, where he was lying and smoking many cigarettes as he usually did, Lord Henry Wotton could see the bright, golden-yellow flowers of a laburnum tree. The laburnum's branches, delicate and trembling, seemed to barely support the weight of their beautiful, flame-like blossoms. Now and then, the dark shapes of flying birds passed quickly across the long, dull-yellow silk curtains in front of the large window. This created a brief, artistic style like Japanese art, reminding Lord Henry of the pale-faced artists from Tokyo. These artists use still images in their art to show a sense of speed and movement. The low, continuous sound of bees pushing through the tall, uncut grass or flying in circles around the dusty, gold-colored stems of the climbing woodbine seemed to make the quiet even more heavy and uncomfortable. | According to the text, what do the flowers, the birds, and the bees have in common?<br><br>a) They all help Lord Henry experience the garden with his senses.<br>b) They all show how beautiful nature is.<br>c) They all make the garden seem busy compared to the stillness of the room.<br>d) They all make Lord Henry think about art and memories.<br>e) They all make Lord Henry think about faraway places. | a | According to the text, what happens when the bees fly through the grass?<br><br>a) The garden seems even more quiet and still.<br>b) The sound of the bees makes the garden feel more alive.<br>c) Lord Henry is distracted by the buzzing sound, which interrupts his thoughts.<br>d) Lord Henry wonders if the bees are attracted to the scent of the blossoms.<br>e) The bees make the blossoms sway in the breeze. | a |
| Lit/Phil | https://www.gutenberg.org/cache/epub/28054/pg28054-images.html | Alexey Fyodorovitch Karamazov was the third son of Fyodor Pavlovitch Karamazov, a land owner well known in our district in his own day, and still remembered among us owing to his gloomy and tragic death, which happened thirteen years ago, and which I shall describe in its proper place. For the present I will only say that this "landowner"—for so we used to call him, although he hardly spent a day of his life on his own estate—was a strange type, yet one pretty frequently to be met with, a type abject and vicious and at the same time senseless. But he was one of those senseless persons who are very well capable of looking after their worldly affairs, and, apparently, after nothing else. Fyodor Pavlovitch, for instance, began with next to nothing; his estate was of the smallest; he ran to dine at other men's tables, and fastened on them as a toady, yet at his death it appeared that he had a hundred thousand roubles in hard cash. At the same time, he was all his life one of the | Alexey Fyodorovitch Karamazov was the third son of Fyodor Pavlovitch Karamazov, a landowner who was well-known in our area during his lifetime. People still remember him because of his dark and tragic death, which happened thirteen years ago. I will describe it later. For now, I'll just say that this ""landowner""—that's what we called him, even though he rarely spent time on his own property—was a strange but common type of person. He was low, mean, and senseless. However, even though he was senseless, he was very good at managing his money and, it seemed, nothing else. For example, Fyodor Pavlovitch started with almost nothing. His land was very small. He would go to other people's houses for dinner and act like a flatterer to get what he wanted. But when he died, it turned out he had one hundred thousand rubles in cash. At the same time, he was one of the most senseless and strange people in the whole area. I repeat, it wasn't stupidity—most of these | Based on the passage, which of the following best describes Fyodor Pavlovitch Karamazov?<br><br>a) A wealthy and respected landowner<br>b) A poor yet generous man<br>c) A brilliant but odd neighbor<br>d) A mean and neglectful father<br>e) A clever but unpleasant person | e | According to the text, what was Fyodor Pavlovitch Karamazov capable of doing very well?<br><br>a) Making friends<br>b) Managing his money<br>c) Raising his sons<br>d) Helping his country<br>e) Tricking other people | b |
| | | most senseless, fantastical fellows in the whole district. I repeat, it was not stupidity—the majority of these fantastical fellows are shrewd and intelligent enough—but just senselessness, and a peculiar national form of it. | strange people are clever and intelligent—but just senselessness, and a particular kind of it that is common in our country. | | | | |
| Lit/Phil | https://www.gutenberg.org/cache/epub/7370/pg7370-images.html | To understand political power right, and derive it from its original, we must consider, what state all men are naturally in, and that is, a state of perfect freedom to order their actions, and dispose of their possessions and persons, as they think fit, within the bounds of the law of nature, without asking leave, or depending upon the will of any other man.<br><br>A state also of equality, wherein all the power and jurisdiction is reciprocal, no one having more than another; there being nothing more evident, than that creatures of the same species and rank, promiscuously born to all the same advantages of nature, and the use of the same faculties, should also be equal one amongst another without subordination or subjection, unless the lord and master of them all should, by any manifest declaration of his will, set one above another, and confer on him, by an evident and clear appointment, an undoubted right to dominion and sovereignty. | To understand political power correctly and where it comes from, we need to think about the natural state of all people. This natural state is one of complete freedom. People can do what they want with their belongings and themselves, as long as they follow the laws of nature. They don't need permission from or rely on anyone else.<br><br>It's also a state of equality, where everyone has the same power and control over others. No one is more powerful than anyone else. It's clear that people of the same kind, born with the same natural abilities and advantages, should be equal. There shouldn't be any hierarchy or control unless a higher power, like a god, clearly states that one person is above another. A god would have to give one person a clear and obvious right to rule and be in charge. | In the author's view, what is the natural state of all human beings?<br><br>a) Nobody is in charge, and everything is confused<br>b) Everyone is free, and everyone is equal<br>c) People have to follow orders and have no real choices<br>d) There is constant fighting and war<br>e) Some people are born better, while others are born worse | b | According to the text, when might one person have the right to rule over another?<br><br>a) When they are stronger than everyone else.<br>b) When they have the most money and land.<br>c) When everyone agrees that they should be the leader.<br>d) When a higher power chooses that person to be the leader.<br>e) When they know more than everyone else. | d |
| Lit/Phil | https://www.gutenberg.org/cache/epub/34901/pg34901-images.html | Such being the reasons which make it imperative that human beings should be free to form opinions, and to express their opinions without reserve; and such the baneful consequences to the intellectual, and through that to the moral nature of man, unless this liberty is either conceded, or asserted in spite of prohibition; let us next examine whether the same reasons do not require that men should be free to act upon their opinions—to carry these out in their lives, without hindrance, either physical or moral, from their fellow-men, so long as it is at their own risk and peril. This last proviso is of course indispensable. No one pretends that actions should be as free as opinions. On the contrary, even opinions lose their immunity, when the circumstances in which they are expressed are such as to constitute their expression a positive instigation to some mischievous act. | Since people must be free to form and share their thoughts without holding back, and since it's harmful to people's minds and morals if this freedom is not allowed or fought for, let's see if the same reasons also mean people should be free to act on their thoughts. They should be able to live their lives based on their beliefs without others physically or morally stopping them, as long as they take the risks themselves. This last part is super important. No one says actions should be as free as thoughts. In fact, even thoughts lose their protection when they're said in a way that encourages someone to do something bad. | According to the passage, why is it important for people to be able to share their opinions freely?<br><br>a) So the government doesn't get too much power.<br>b) So people can learn and grow, intellectually and morally.<br>c) So people can find out what's really true.<br>d) So everyone can agree on things.<br>e) So people can have new ideas and get richer. | b | Under what circumstances does the author believe that people should NOT be allowed to freely express their opinions?<br><br>a) When their opinions are unpopular<br>b) When their opinions might lead to violence or other harmful actions<br>c) When their opinions are not based on facts<br>d) When their opinions are offensive to others<br>e) The author does not believe that there should be any limits on free speech | b |
| Lit/Phil | https://www.gutenberg.org/cache/epub/147/pg147-images.html | Society in every state is a blessing, but government even in its best state is but a necessary evil; in its worst state an intolerable one; for when we suffer, or are exposed to the same miseries by a government, which we might expect in a country without government, our calamity is heightened by reflecting that we furnish the means by which we suffer. Government, like dress, is the badge of lost innocence; the palaces of kings are built on the ruins of the bowers of paradise. For were the impulses of conscience clear, uniform, and irresistibly obeyed, man would need no other lawgiver; but that not being the case, he finds it necessary to surrender up a part of his property to furnish means for the protection of the rest; and this he is induced to do by the same prudence which in every other case advises him out of two evils to choose the least. Wherefore, security being the true design and end of government, it unanswerably follows that whatever form thereof appears most likely to ensure it to us, with the least expence and greatest benefit, is preferable to all others. | Society in any form is a good thing, but government, even when it's good, is still a necessary problem. When it's bad, it's unbearable. When we suffer under a government the same way we would without one, it's even worse because we know we're causing our own problems by supporting it. Government, like clothing, shows that we've lost our innocence. Royal palaces are built where beautiful, natural places used to be.<br><br>If our conscience, our inner sense of right and wrong, was always clear and strong, we wouldn't need rules. But since it's not, we give up some of our belongings to pay for protection of the rest. We do this because we're smart and know that, when faced with two bad choices, we should pick the less bad one.<br><br>Since the real purpose of government is to keep us safe, it's clear that the best form of government is the one that keeps us safe the best, costs the least, and benefits us the most. | According to the text, what is the primary reason why government is necessary?<br><br>a) To protect citizens from outside threats.<br>b) To secure political and economic justice.<br>c) To ensure the fair distribution of wealth.<br>d) To compensate for the imperfections of human nature.<br>e) To promote economic growth. | d | According to the passage, what does the author think about government and society?<br><br>a) Government is really important for society, and it does many good things.<br>b) Government is bad, but we need it.<br>c) Government stops people from being free, so we should have as little as possible.<br>d) Government shows that people are naturally good.<br>e) A higher power wants us to have a government. | b |
| Lit/Phil | https://www.gutenberg.org/cache/epub/3300/pg3300-images.html#chap02 | The greatest improvements in the productive powers of labour, and the greater part of the skill, dexterity, and judgment, with which it is anywhere directed, or applied, seem to have been the effects of the division of labour. The effects of the division of labour, in the general business of society, will be more easily understood, by considering in what manner it operates in some particular manufactures. It is commonly supposed to be carried furthest in some very trifling ones; not perhaps that it really is carried further in them than in others of more importance: but in those trifling manufactures which are destined to supply the small wants of but a small number of people, the whole number of workmen must necessarily be small; and those employed in every different branch of the work can often be collected into the same workhouse, and placed at once under the view of the spectator. | The biggest improvements in how much work people can do, and most of the skill and knowledge used in work, seem to be because of the division of labor, a way of organizing work. It's easier to understand how the division of labor affects society by looking at how it works in specific industries. It's often thought to be used the most in small, unimportant industries. This isn't necessarily because it's used more there than in other, more important industries. It's because these small industries, which make things for only a few people, have a small number of workers. The workers in each part of the process can often be put in the same building, where someone can easily see them all at once. | According to the passage, how does the author feel about dividing up work?<br><br>a) It's a good idea, but only for simple jobs.<br>b) It's a good idea, because it helps everyone get more done.<br>c) It's a bad idea, because it makes work boring for people.<br>d) It's a bad idea, because it makes it harder for people to learn new skills.<br>e) It doesn't really matter whether work is divided up or not. | b | Why does the author think it's easier to see how dividing up work helps in smaller businesses?<br><br>a) Workers in small businesses are better at their jobs.<br>b) Small businesses do easier work.<br>c) In small businesses, you can see all the workers doing their different jobs together.<br>d) There is a larger number of small businesses, so it's easier to find examples.<br>e) Workers in small businesses get along better. | c |
| Aero/CS | https://science.nasa.gov/mission/voyager/interstellar-mission/ | Voyager 1 entered interstellar space on Aug. 25, 2012, becoming the first human-made object to do so. Voyager 1 entered interstellar space at about 122 AU, or about 11 billion miles (18 billion kilometers) from the Sun. However, it was not immediately clear to the Voyager science team that Voyager 1 had crossed the heliopause. Particle instruments on the spacecraft showed an increase in cosmic rays (which originate outside the heliosphere) and decreases in heliospheric particles. But what would have been an additional source of confirmation, the probe's plasma science instrument (PLS), had stopped working in 1980. The PLS was designed to measure the speed and direction of the solar wind while Voyager 1 was inside the heliosphere, and in interstellar space, it would | Voyager 1, the first human-made object to enter interstellar space, crossed into it on August 25, 2012. Voyager 1 was about 122 AU, or roughly 11 billion miles (18 billion kilometers) from the Sun when it entered interstellar space. However, the Voyager science team wasn't sure right away that Voyager 1 had crossed the heliopause, the boundary between the heliosphere and interstellar space. Instruments on the spacecraft showed more cosmic rays, which come from outside the heliosphere, and fewer particles from inside the heliosphere. But the probe's plasma science instrument (PLS), which could have provided more proof, stopped working in 1980. The PLS was built to measure the speed and direction of the solar wind while Voyager 1 was inside the | What was the main challenge the Voyager science team faced in confirming that Voyager 1 had entered interstellar space?<br><br>a) The spacecraft was too far away to communicate with effectively.<br>b) The spacecraft's instruments were damaged by the harsh conditions of interstellar space.<br>c) The spacecraft's primary instrument for measuring the solar wind had stopped working.<br>d) The scientists were unsure of the exact location of | c | What is the heliopause, as described in the passage?<br><br>a) The boundary between the solar system and interstellar space.<br>b) The region of space where the solar wind is strongest.<br>c) The point where the Sun's gravity no longer has any effect.<br>d) The area where cosmic rays are most intense.<br>e) The center of the Milky Way galaxy. | a |
| | | have detected a dramatic drop in those measurements. Without the plasma science instrument, the Voyager science team couldn't be sure the probe had left the heliosphere. Thankfully, almost a year later, Voyager 1's plasma wave subsystem (PWS) detected oscillations in the plasma surrounding the spacecraft, which indicated that Voyager 1 was moving into an increasingly dense region of plasma, a clear indicator that the probe had entered interstellar space. | heliosphere. In interstellar space, it would have shown a big drop in those measurements. Without the PLS, the Voyager science team couldn't be certain the probe had left the heliosphere. Luckily, almost a year later, Voyager 1's plasma wave subsystem (PWS) found changes in the plasma around the spacecraft. These changes showed that Voyager 1 was going into a denser area of plasma, a clear sign that the probe had entered interstellar space. | the heliopause.<br>e) The data from the spacecraft was difficult to interpret and understand. | | | |
| Aero/CS | https://en.wikinews.org/wiki/NASA_issues_survivability_report_on_Columbia_crash | The report notes that the cabin began spinning wildly, causing some of the astronauts' helmets to come off. The report noted that most on board were secured only by lap harnesses that offered no restraint to the upper body and were not designed to cope with sideways motion, meaning "lethal trauma" was caused by the rotation. However, the report was unable to determine whether the astronauts had died from oxygen deprivation or the extreme nature of their injuries.<br><br>The pressure suits the crew wore, introduced after the Space Shuttle Challenger disaster, were criticised as the crew could not keep their visors down throughout re-entry due to a design limitation that would have caused excess levels of oxygen to be present had they done so, leaving the suits unsealed. The gloves also made many tasks difficult or impossible. When the accident occurred, three people were not wearing gloves, one was not wearing a helmet, and none had their visors down. The helmets also did not conform to the heads of the wearers. | The report says the cabin started spinning very fast, making some astronauts' helmets fall off. Most astronauts were only secured by lap belts, which don't protect the upper body and aren't meant for side-to-side movement. This spinning caused severe injuries that could have been deadly. But the report couldn't say if the astronauts died from not getting enough oxygen or from the bad injuries.<br><br>The pressure suits, which were used after the Space Shuttle Challenger disaster, were criticized. The astronauts couldn't keep their face shields down during re-entry because of a design problem. If they had, too much oxygen would have been in the suits, and they wouldn't have been sealed properly. The gloves also made it hard or impossible to do many tasks. When the accident happened, three people weren't wearing gloves, one wasn't wearing a helmet, and no one had their face shields down. The helmets also didn't fit the astronauts' heads well. | Why could the Space Shuttle crew not keep their visors down?<br><br>a) They needed oxygen from the shuttle to come in.<br>b) The cabin was spinning too much.<br>c) The gloves made it hard to close the visor.<br>d) The crew were not wearing gloves when they should have been.<br>e) The visors were not designed well. | e | What was wrong with the harnesses?<br><br>a) They did not secure the helmet.<br>b) They secured the lap.<br>c) They did not provide oxygen.<br>d) They did not secure against rotation.<br>e) They were hard to use with gloves on. | d |
| Aero/CS | https://en.wikinews.org/wiki/Turing_test_beaten_by_Russian_chatterbot | The Turing test is a test of artificial intelligence aiming to fulfil the suggestion of Alan Turing in his 1950 paper "Computing Machinery and Intelligence", which stated that within fifty years, an "average interrogator" would, following a five-minute long conversation, "not have more than 70 per cent chance" of correctly predicting whether they are speaking to a human, or a machine — which would be able to, as such, fool at least 30% of human judges into thinking it is human.<br><br>In the contest, where Eugene Goostman and four other bots competed, the bot successfully tricked 33% of the participating judges, which included television actor Robert Llewellyn of the BBC television series Red Dwarf, and John Sharkey, Baron Sharkey, a sponsor of Turing's 2013 posthumous pardon. To give the bot a "believable personality", Goostman is portrayed as being a thirteen year-old boy of the Ukraine; the bot's head developer Vladimir Veselov stated that this made Goostman "not too old to know everything and not too young to know nothing". | The Turing test checks if a computer can think like a human. Alan Turing, in his 1950 paper "Computing Machinery and Intelligence," suggested this test. He thought that in fifty years, a person asking questions would not be able to tell if they were talking to a human or a machine more than 70% of the time after a five-minute conversation. This means the machine would trick at least 30% of people into thinking it was human.<br><br>In a competition, Eugene Goostman and four other computer programs competed. Goostman tricked 33% of the judges, including Robert Llewellyn, a TV actor from the show Red Dwarf, and John Sharkey, Baron Sharkey, who supported Turing's pardon after his death in 2013. To make Goostman seem more real, it was designed to be a thirteen-year-old boy from Ukraine. Vladimir Veselov, the bot's main developer, said this made Goostman believable because he was "not too old to know everything and not too young to know nothing." | What is the Turing test?<br><br>a) To test if a person is as good as an artificial intelligence.<br>b) To test if an artificial intelligence can have a conversation.<br>c) To test if artificial intelligence can act as a judge.<br>d) To test if an artificial intelligence can converse as well as a human.<br>e) To test if an artificial intelligence can converse well enough that human judges think they are talking to a human. | e | Why was the artificial intelligence designed to be a 13 year old from Ukraine?<br><br>a) So that the artificial intelligence can be illogical at times.<br>b) So that the artificial intelligence did not need to handle hard questions.<br>c) So that the artificial intelligence can provide playful instead of factual responses.<br>d) So that the artificial intelligence can respond slowly.<br>e) So that the artificial intelligence can provide childish responses without giving it away. | b |
| Aero/CS | https://en.wikinews.org/wiki/Welsh_University_announces_intelligent_robot_conducting_biology_experiments | Adam is a robot developed by Welsh Aberystwyth University researchers which combines artificial intelligence (AI), robotics and automation to independently conduct and analyse biological laboratory research.<br><br>Adam has spawned discussion amongst researchers. William Melek, an assistant professor of mechanical engineering at Ontario's University of Waterloo, has noted that to set up the AI needed for subsequent experiments involving new biological variables and criteria, the human expertise would be time consuming to customize it. The usefulness would be limited therefore to the allotment of human input needed to set up Adam.<br><br>David Waltz of Columbia University and Bruce Buchanan of the University of Pittsburgh say that "For the foreseeable future, the prospect of using automated systems as assistants holds vast promise as these assistants are becoming not only faster but much broader in their capabilities -- more knowledgeable, more creative, and more self-reflective," They note the potential of such lab assistants which may more efficiently process the research data. | Adam is a robot created by scientists at Aberystwyth University in Wales. It uses artificial intelligence (AI), robotics, and automation to do and analyze biological research in a lab without needing a person to help.<br><br>Adam has caused researchers to talk about it. William Melek, a teacher of mechanical engineering at the University of Waterloo in Ontario, said that to set up the AI for new experiments, scientists would need to spend a lot of time customizing it. This is because the AI needs new information about biological things and the goals of the experiment. So, Adam's usefulness would be limited by how much time people need to set it up.<br><br>David Waltz from Columbia University and Bruce Buchanan from the University of Pittsburgh said that, "In the near future, using automated systems as helpers is very promising because these helpers are getting faster and better at more things. They are becoming more knowledgeable, more creative, and more able to think about themselves." They also mentioned that these lab helpers could process research data more quickly. | Why might Adam, the robot's usefulness be limited?<br><br>a) It is too hard to make another Adam.<br>b) It is too expensive to make another Adam.<br>c) It is too time confusing to customize Adam.<br>d) Adam is too fast and jumps to conclusions.<br>e) Adam is too creative and therefore inaccurate. | c | What might automated systems such as Adam be promising as a lab assistant?<br><br>a) They can set themselves up.<br>b) They get cheaper over time.<br>c) They can provide the right answer to an experiment.<br>d) They get more capable over time.<br>e) They can teach humans. | d |